\documentclass{article}

\PassOptionsToPackage{numbers,compress}{natbib}
\usepackage[preprint]{neurips_2026}

\usepackage[utf8]{inputenc}
\usepackage[T1]{fontenc}
\usepackage{hyperref}
\usepackage{url}
\usepackage{booktabs}
\usepackage{amsfonts}
\usepackage{amsmath}
\usepackage{array}
\usepackage{longtable}
\usepackage{multirow}
\usepackage{calc}
\usepackage{graphicx}
\usepackage{caption}
\usepackage{fvextra}
\usepackage{nicefrac}
\usepackage{microtype}
\usepackage[table]{xcolor}

\providecommand{\tightlist}{%
  \setlength{\itemsep}{0pt}\setlength{\parskip}{0pt}}
\providecommand{\rev}[1]{#1}
\providecommand{\revblock}[1]{#1}

\title{SEEK-VAU: Towards Evidence-Faithful Video Anomaly Understanding via Agentic Search}

\author{%
  Minghe Wang$^{1}$ \quad Weijun Zeng$^{3}$ \quad Dapeng Luo$^{4}$ \quad Xinyue Wang$^{4}$ \quad Nong Sang$^{2}$\thanks{Corresponding author} \\
  $^{1}$ School of Software Engineering, Huazhong University of Science and Technology \\
  $^{2}$Key Laboratory of Image Processing and Intelligent Control, \\
  School of Artificial Intelligence and Automation, Huazhong University of Science and Technology \\
  $^{3}$School of Software \& Microelectronics, Peking University \\
  $^{4}$School of Mechanical Engineering and Electronic Information, China University of Geosciences \\
  \texttt{\{minghe\_wang, nsang\}@hust.edu.cn} \quad \texttt{zengweijun26@stu.pku.edu.cn} \\
  \texttt{\{luodapeng, wangxinyue\}@cug.edu.cn}
}

\hypersetup{
  pdftitle={SEEK-VAU: Towards Evidence-Faithful Video Anomaly Understanding via Agentic Search},
  pdfauthor={Minghe Wang, Weijun Zeng, Dapeng Luo, Xinyue Wang, Nong Sang},
  pdfkeywords={video anomaly understanding, agentic search, evidence verification, multimodal large language model, reinforcement learning}
}

\begin{document}

\maketitle

\begin{abstract}
Most video anomaly understanding (VAU) systems still infer anomaly
judgments from evidence collected in a feed-forward way under
predefined rules, typically sampled frames, or events clustered from such frames.
This protocol fails when the collected evidence is
incomplete under imperfect predefined rules, because once reasoning
begins the model can no longer search for the missing evidence.
We address this limitation with \textbf{SEEK-VAU}, which reframes VAU
as an \textbf{agentic search-and-verify interaction protocol} under a
bounded search budget:
instead of passively reasoning over pre-collected evidence, the policy
actively gathers evidence across the three stages of an anomaly event
chain (precursor, trigger, and aftermath) and verifies its sufficiency
before finalization, turning event-chain completeness into an explicit
design rather than an outcome of predefined rules.
A bounded budget caps the cost of this search,
but does not by itself teach the policy where to look or when to stop.
To make this behavior learnable, we introduce 
\textbf{evidence-faithful counterfactual verification (EFCV)}, which
rewards selected evidence that remains supportive, compact, and
necessary under counterfactual verification. We further introduce
\textbf{SEEK-Bench}, featuring video-level episodes with temporal
interval annotations, semantic QA and event-chain stage labels. Together,
SEEK-VAU and SEEK-Bench establish a strong foundation for evidence-faithful
and actively verifiable VAU. 
\end{abstract}

\hypertarget{introduction}{%
\section{Introduction}\label{introduction}}

Video anomaly analysis has moved from detecting and localizing abnormal
events toward explaining what happened, why it is anomalous, and what
visual support justifies the conclusion \citep{du2024uncovering,yang2024follow,tang2024hawk,zhang2025holmes,ye2025vera,xu2025towards} (Figure \ref{fig:teaser}(a)).
This shift has pushed VAU toward longer temporal reasoning, richer
context modeling, and more structured decision making,
including methods that operate across long videos, multiple temporal granularities,
subtle scene and reinforcement-enhanced reasoning
\citep{zhang2025holmes,zhu2025vau,huang2026advancing,yu2026cuebench} (Figure \ref{fig:teaser}(b)). Yet the dominant protocol remains
fixed-observation: a feed-forward pipeline driven by imperfect predefined rules, the model receives a prepared bundle of frames or events and then infers an anomaly judgment, explanation, or
reasoning from that bundle. Stronger reasoning can improve
interpretation over observed evidence, but it does not change when or
how evidence is acquired, leaving the model unable to search for missing
support once inference has begun.

\begin{center}
\captionsetup{hypcap=false}
\includegraphics[width=0.8\textwidth]{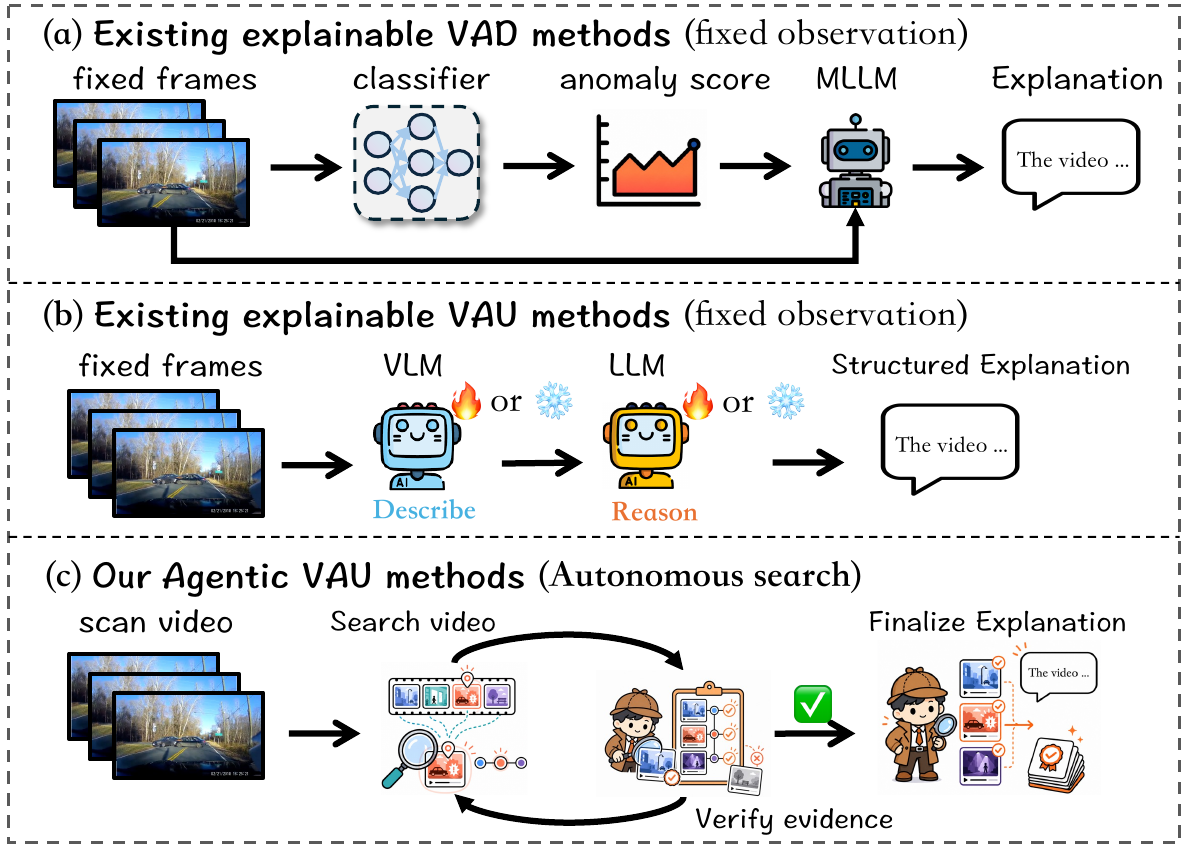}
\captionof{figure}{(a) Explainable VAD and (b) reasoning-enhanced VAU explain anomalies from a fixed observation bundle preselected before reasoning, so missed evidence can never be recovered. (c) \mbox{SEEK-VAU} interleaves acquisition with reasoning: it scans, iteratively seeks missing event-chain evidence in a seek--verify loop, and finalizes only when the gathered evidence supports the case, under a bounded search budget.}
\label{fig:teaser}
\end{center}

This interaction-level limitation matters because many anomalies are not
defined by a single peak frame or short clip, but by whether the system
can recover a coherent chain from precursor cues to a
trigger, and then to aftermath. Under this lens of
\textbf{event-chain completeness}, fixed-observation VAU becomes
structurally limiting: when key evidence is missing, the model must 
reason over an incomplete observation set rather than search for missing
evidence or verify the case's readiness for
closure. Recent agentic systems PANDA
\citep{yang2025panda} show that multi-round tool-augmented
reflection can recover useful evidence, but they cap only
reflection rounds without bounding per-turn or cumulative context, so inference cost grows with accumulating outputs.
SEEK-VAU addresses this gap by reframing VAU as an
\textbf{agentic search-and-verify} problem under a
bounded search budget, in which the policy acquires evidence,
verifies its readiness, and commits to a final anomaly
explanation.

Concretely, \textbf{SEEK-VAU} instantiates this search-and-verify view
as a tool-using policy, following the reasoning-action-observation paradigm
\citep{yao2022react}, organized around four executable actions:
$\{\texttt{scan}, \texttt{seek}, \texttt{verify}, \texttt{finalize}\}$.  
\texttt{scan} builds an initial
temporal context over the video, whereas \texttt{seek} uses that
context to turn promising regions into committed supporting evidence;
\texttt{verify} then tests whether the selected evidence is sufficient, insufficient or redundant before \texttt{finalize} closes the case.
This protocol is more than a cleaner interface over existing VAU, as its decomposition allows the policy to search backward for precursors or forward for aftermath, rather than remaining trapped inside a fixed observation bundle chosen before reasoning begins.
Without a cost
mechanism, however, this freedom would inflate inference cost as the
policy revisits the video. SEEK-VAU prevents this with a bounded
search budget that imposes two coordinated caps: a per-turn cap on
the frame intake of each \texttt{seek}, and a
cumulative cap maintained by a rolling active-context window that
evicts older frames. Per-step compute therefore stays
bounded as turns accumulate, so backtracking improves recovery
without significantly inflating inference cost.

Protocol alone, however, does not guarantee that the policy will
actually learn to search and stop for evidence-faithful reasons. We
therefore pair SEEK-VAU with a learning objective grounded in
\textbf{evidence-faithful counterfactual verification (EFCV)}. The supervised fine-tuning (SFT) stage trains the policy on prepared protocol
trajectories that instantiate the same search-verify-finalize interface used
at inference, and the reinforcement learning (RL) stage rewards anomaly
trajectories whose committed evidence supports the decision, whose
compact subset remains sufficient, and whose trigger evidence is
decision-critical. Evidence faithfulness thus becomes part of policy
optimization itself, rather than a post-hoc aspiration.

To evaluate whether search-and-verify behavior improves evidence
recovery rather than only final answers, we introduce
\textbf{SEEK-Bench}. To our knowledge, it is the first benchmark
designed for interactive search-and-verify VAU, annotating each episode
with temporal intervals, semantic QA and event-chain stages. To this end, we make three contributions:

\vspace{-0.4em}
\begin{itemize}
\tightlist
\item
  We recast VAU as a stage-aware \textbf{agentic search-and-verify protocol
under a bounded search budget}, and instantiate it
in SEEK-VAU through four executable actions that make event-chain
completeness a recoverable behavioral target rather than a passive
byproduct of fixed observation, while the budget
bounds per-turn input tokens and cumulative visual context so
backtracking improves recovery without significantly inflating
inference cost.
\item
  We introduce \textbf{EFCV}, which couples
  protocol-trajectory supervision with counterfactual verification rewards
  for selected-support sufficiency, compactness, and trigger necessity,
  training final decisions to rely on committed evidence rather than
  label correctness alone.
\item
  We introduce \textbf{SEEK-Bench}, a benchmark
  for interactive search-and-verify VAU, with video episodes
  annotated for temporal intervals, semantic QA and event-chain
  stages.
\end{itemize}
\vspace{-0.7em}

\hypertarget{related-work}{%
\section{Related work}\label{related-work}}

\hypertarget{from-video-anomaly-detection-to-richer-vau}{%
\subsection{From video anomaly detection to richer
VAU}\label{from-video-anomaly-detection-to-richer-vau}}

Video anomaly research has shifted from detecting and localizing
abnormal events to explaining what happened, why it is anomalous, and
how it unfolds. CUVA \citep{du2024uncovering} and HAWK \citep{tang2024hawk} emphasize causation-oriented and open-world understanding, Holmes-VAU \citep{zhang2025holmes} extends
VAU to long videos and multiple temporal granularities, and CueBench \citep{yu2026cuebench}
highlights context-dependent anomalies that require scene- and
attribute-aware interpretation. Prompting-based systems such as
AnomalyRuler \citep{yang2024follow}, VERA \citep{ye2025vera}, and AssistPDA \citep{yang2025assistpda} further enrich anomaly description and
explanation. Together, these works broaden VAU's semantic
scope, but the resulting pipelines still largely reason over prepared
frames or hierarchical events, leaving the fixed-observation protocol unchanged.

\hypertarget{reasoning--and-rl-enhanced-vau}{%
\subsection{Reasoning- and RL-enhanced
VAU}\label{reasoning--and-rl-enhanced-vau}}

Recent VAU methods strengthen reasoning over provided video evidence.
VAU-R1 \citep{zhu2025vau} organizes perception, grounding, reasoning, and conclusion with
reinforcement fine-tuning; SRVAU-R1 \citep{zhao2026srvau} and PrismVAU \citep{erregue2026prismvau} add reflection-aware learning and prompt-refined multimodal inference; Vad-R1 \citep{huang2025vad} and VADER \citep{cheng2026vader} emphasize chain-of-thought and causal
interpretation; and Huang et al. \citep{huang2026advancing} introduce adaptive
hierarchical reasoning, risk-aware analysis, and decision-oriented
outputs. These methods make VAU more structured and adaptive,
but they remain fixed-observation systems: they reason over provided
video-question context and cannot actively search for missing precursor,
trigger, or aftermath evidence needed to reconstruct complete anomaly
event chains.

\hypertarget{agentic-video-anomaly-detection}{%
\subsection{Agentic Video Anomaly Detection}\label{agentic-video-anomaly-detection}}

Beyond fixed-observation VAU, recent work casts video anomaly
detection (VAD) itself as an agentic process. PANDA \citep{yang2025panda} frames generalized anomaly analysis as a training-free VAD agent with planning, tool use, memory,
and external evidence tools such as object detection and web search. QVAD \citep{bekit2026qvad} remains detection-centric but uses
question-driven inquiry and iterative query refinement over the provided
visual input. These systems show that anomaly inference can be
interactive, yet they do not instantiate agentic VAU: PANDA operationally supports iterative reasoning, but it does not evaluate stage-wise evidence recovery under a matched cumulative visual-context budget, while QVAD
remains fixed-input interactive reasoning rather than
tool-mediated evidence acquisition. 

Taken together, prior VAU work improves semantic understanding and
reasoning, while agentic VAD introduces interactive inference. 
Unlike general agentic VAD agents,
SEEK-VAU makes anomaly event-chain evidence the unit of search, verification,
optimization, and evaluation. SEEK-VAU closes this gap with an agentic
search-and-verify protocol under a bounded search budget.
Counterfactual signals have appeared in image-grounded reasoning
(DeFacto~\citep{defacto2025}) and test-time-scaled video reasoning
(Video-RTS~\citep{videorts2025}), but neither targets event-chain
anomaly understanding; EFCV ties counterfactual perturbations to
stage-tagged evidence (precursor, trigger, aftermath), isolating
trigger-specific necessity rather than generic evidence sufficiency.

\hypertarget{seek-vau-method}{%
\section{SEEK-VAU: method}\label{seek-vau-method}}

\hypertarget{task formulation}{%
\subsection{Task formulation}\label{task-formulation}}

Most VAU systems answer from a fixed observation bundle selected before
inference, which is limiting when anomaly meaning depends on evidence
outside that bundle. Given a video \(V\) and a case-level task query
\(Q\), SEEK-VAU frames VAU as a search-and-verify task that returns a
structured anomaly case \(y\), including anomaly existence, category, a
grounded interval, supporting evidence, and a semantic explanation,
rather than a one-shot judgment over precommitted observations.

The central target of this interaction is a coherent anomaly event chain
\(C=(C_{\mathrm{pre}}, C_{\mathrm{trg}}, C_{\mathrm{aft}})\), where the
three components denote evidence recovered for precursor, trigger, and
aftermath stages when such stages are observable. Some cases may lack a
visible precursor or aftermath, so the objective is not to force every
episode into a complete three-stage template. Instead, under a bounded
search budget, the policy should recover as much of the relevant event
chain as possible and use the selected evidence to produce an
evidence-faithful reasoning result.

\begin{center}
\captionsetup{hypcap=false}
\includegraphics[width=\textwidth]{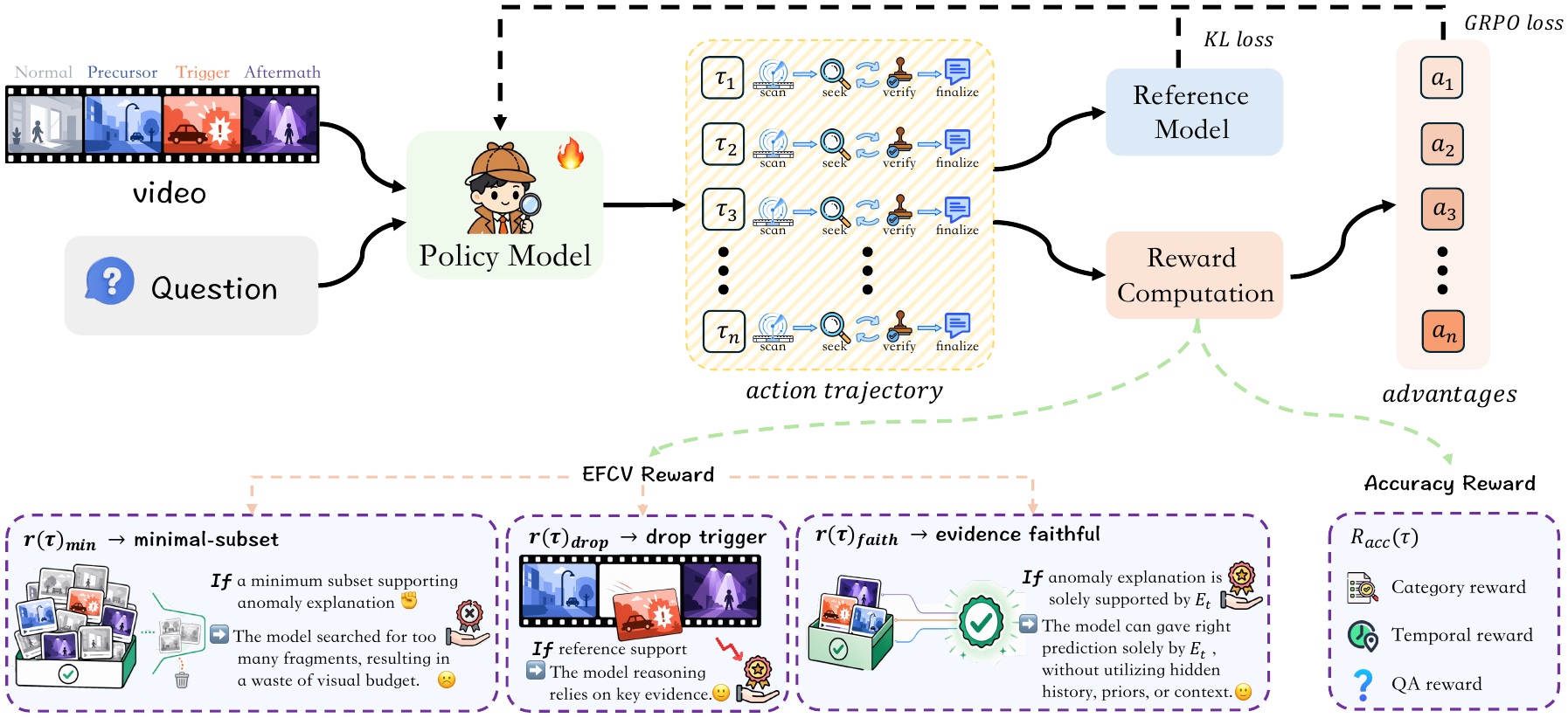}
\captionof{figure}{Overview of our proposed SEEK-VAU}
\label{fig:method-overview}
\end{center}

\hypertarget{interleaved-event-chain-search}{%
\subsection{Interleaved event-chain
search}\label{interleaved-event-chain-search}}

Fixed-observation VAU separates evidence acquisition from reasoning, as
the visual bundle is chosen before inference and remains unchanged while
the anomaly hypothesis evolves. This is limiting
for event-chain understanding: when a reasoning step exposes a missing
stage, the model cannot search for the evidence needed to resolve it. SEEK-VAU therefore treats
search as an interleaved evidence
acquisition process inside anomaly reasoning. The goal is not simply to
inspect more frames, but to recover the missing stage
needed for evidence-faithful closure under a bounded search budget.

At turn \(t\), the rollout state is summarized as
\(h_t=(M_t,E_t,c_t,H_t, Q)\), where \(M_t\) is the temporal search map,
\(E_t\) is the dynamic evidence set, \(c_t\) is the working anomaly
hypothesis, \(H_t\) is the interaction history, and \(Q\)
is the case-level task query, which is fixed throughout the episode. The current event-chain coverage is induced from the
dynamic evidence set as \(C_t=\Psi(E_t)\), so search progress is measured
by which precursor, trigger, and aftermath evidence has actually been
committed rather than by how many frames have been viewed. Among the four protocol actions
\(\{\texttt{scan}, \texttt{seek}, \texttt{verify}, \texttt{finalize}\}\),
\texttt{scan} and \texttt{seek} are responsible for updating the temporal search map \(M_t\) and the
dynamic evidence set \(E_t\). To make this separation explicit, we decompose the interaction into
evidence acquisition and evidence-conditioned finalization:

\begin{equation}
P_\pi(y, E_T \mid V, Q) =
\underbrace{P_\pi(E_T \mid V, Q)}_{\text{evidence acquisition}}
\cdot
\underbrace{P_\pi(y \mid E_T, V, Q)}_{\text{case finalization}}
\end{equation}

Here \(E_T\) denotes the dynamic evidence set accumulated by the end of
the interaction. The first factor corresponds to evidence acquisition:
through the first \texttt{scan} and repeated \texttt{seek} calls, the policy builds
temporal context, then emits local queries \(q_t\),
and commits selected evidence. The second factor
corresponds to finalization: the policy predicts the structured anomaly
case \(y\) conditioned on the gathered evidence rather than on a fixed
observation bundle selected before reasoning begins.

SEEK-VAU implements interleaved evidence search through its two
evidence-acquisition actions. \texttt{scan} explores a requested interval and logs the returned frames
and their timestamps in the temporal search map \(M_t\), a coarse guide
over observed intervals that informs later decisions without itself
becoming selected support. \texttt{seek} converts the current temporal context into
committed evidence: conditioned on the current state
\(h_t\), the policy emits a search
request \(a_t=(q_t, I_t, \rho_t)\), consisting of a concrete visual query
\(q_t\), a temporal interval \(I_t\) in seconds, and the event-chain role
\(\rho_t\) being sought, and commits the returned frames to \(E_t\).

The environment executes each request \(a_t\) through a fixed,
query-independent mapping: the requested interval \(I_t\) is clipped
to the video span, and the tool deterministically returns four
uniformly spaced frames over the clipped interval, together with their
timestamps. Within-window sampling is therefore deterministic and
independent of the query \(q_t\): the query and the event-chain role
\(\rho_t\) record the policy's targeted evidence need, while the
learned decision is which temporal window to inspect.
Each return is committed as one evidence window, the interval
from its earliest to its latest returned frame. The tool
consumes only the input video and the policy-issued interval \(I_t\); it
holds nothing unavailable to a fixed-observation baseline, and learning
where to look reduces to choosing \(I_t\).

The seek query \(q_t\) operationalizes
the policy's current evidence need: it specifies the concrete visual
event, scene, or stage cue that should be searched next, rather than
serving as a post-hoc explanation. SFT demonstrates
this action space with prepared trajectories; RL rollouts and inference
generate \(q_t\) online with the identical procedure, differing only in
the learning signal.

Figure~\ref{fig:teaser}(c) traces one episode: a scan
first establishes temporal context, \texttt{seek} calls commit localized
evidence at stages flagged by the temporal map, and verification
redirects later searches toward missing event-chain stages
before finalization, closing the interaction into a
\texttt{seek}--\texttt{verify} loop. To bound this multi-turn search within a compact visual context, SEEK-VAU maintains
a visible 16-frame active context rather than an unbounded history: the initial 6-frame full-video scan
is protected as a global temporal map; each later seek returns 4 frames; and when admitting a seek result
would exceed the 16-frame cap, the oldest non-protected seek frames are evicted first.

\hypertarget{policy-internal-evidence-verification}{%
\subsection{Policy-internal evidence
verification}\label{policy-internal-evidence-verification}}

\texttt{seek} alone can retrieve relevant windows without
ensuring that the event chain is complete: the policy might commit a
window for the trigger event but never commit any window for a precursor
or an aftermath, in which case calling \texttt{finalize} would close the
case on partial evidence. SEEK-VAU therefore inserts a self-verification
step before \texttt{finalize}, in which the policy reviews its own
evidence set and decides whether the case is ready to be closed.

Given a working hypothesis \(c_t\) and selected evidence \(E_t\),
the policy self-verifier labels the set's support as sufficient, insufficient, or redundant by checking whether the
committed support covers the anomaly stages required by the current
case and avoids unnecessary redundancy. The result is a control
diagnosis of what the current \(E_t\) can justify and what remains missing
before finalization. At turn \(t\), we write the self-verification
interface as the following action contract:

\begin{equation}
\mathrm{Verify}(c_t,E_t)=\bigl(v_t,\, a_t,\, s_{\mathrm{anom}}(c_t,E_t),\, \Delta_t\bigr).
\end{equation}

Here, {\(v_t \in \{\texttt{sufficient}, \texttt{insufficient}, \texttt{redundant}\}\) is the categorical verdict that labels the support state of the committed evidence, and} \(a_t\) recommends whether to continue with \texttt{seek} or move toward
\texttt{finalize}.
\(s_{\mathrm{anom}}(c_t,E_t)\in[0,1]\) is the anomaly
confidence that the policy itself assigns to the current
(claim, evidence) pair.
\(\Delta_t\) summarizes missing stages to resolve before closure.

These outputs make verification an online control gate rather than a
post-hoc score: instead of rescoring a completed answer, it reshapes
the remaining trajectory. When the case is not yet ready, \(\Delta_t\) becomes
the next search target and \(a_t\) routes the policy back to \texttt{seek} within the remaining search budget to recover
the missing precursor, trigger, or aftermath stage. As long as \(v_t\)
reports an unsupported state, \(\Delta_t\) and \(a_t\) redirect the
policy toward the missing stage, and the
policy proceeds to \texttt{finalize} only once \(v_t\) reports sufficient. 
Each such reroute draws on the same search budget, so verification-directed backtracking improves coverage without inflating visual cost.

\hypertarget{efcv-learning}{%
\subsection{EFCV learning}\label{efcv-learning}}

The search-and-verify protocol defines what the policy can do, but
learning still needs to decide which anomaly trajectories are
evidence-faithful. A terminal answer reward can encourage correct labels
and plausible temporal grounding, but it cannot tell whether the
selected evidence actually supports the anomaly decision, whether
redundant windows hide weak support, or whether trigger evidence is
decision-critical. EFCV addresses this gap by converting verifier
diagnostics into learning-time counterfactual pressure:
rather than rewarding a correct answer alone, it rewards
trajectories whose final claim remains justified by the evidence the
policy actually selected. EFCV thus ties evidence faithfulness to budget thrift, rewarding compact, sufficient evidence subsets.

Formally, the rollout-level reward combines answer correctness with
counterfactual verification:

\vspace{-0.5em}
\begin{equation}
R(\tau)
= w_{\mathrm{acc}} R_{\mathrm{acc}}(\tau) +
w_{\mathrm{efcv}}\bigl(
r_{\mathrm{faith}}(\tau)+r_{\mathrm{min}}(\tau)+r_{\mathrm{drop}}(\tau)
\bigr),
\end{equation}
\vspace{-0.5em}

Here, \(w_{\mathrm{acc}}\) and \(w_{\mathrm{efcv}}\)
balance answer correctness and counterfactual verification. 
\(R_{\mathrm{acc}}\) rewards output correctness by aggregating task-level scores for anomaly category, temporal
localization, and the episode's semantic QA, encouraging the
policy to produce a structurally correct and semantically aligned case
report. By contrast, \(R_{\mathrm{efcv}}\) evaluates the evidential
basis of that decision under counterfactual perturbations. It tests
whether the selected evidence can sustain the anomaly claim, whether the
support admits a compact sufficient subset, and whether the trigger
evidence is necessary for the decision. We single out trigger necessity
because the trigger marks the decisive transition from normal context to
anomalous behavior, so a faithful evidence set should weaken support when
this segment is removed. As shown in Figure~\ref{fig:method-overview}, \(R_{\mathrm{efcv}}\)
is computed from three rewards defined below.

\vspace{-0.5em}
\begin{equation}
\begin{array}{@{}l@{\;}c@{\;}l@{}}
r_{\mathrm{faith}}(\tau) &=& \mathbf{1}\!\left[s_{\mathrm{anom}}(c_T,E_T)\ge\theta\right],\\
r_{\mathrm{min}}(\tau) &=& s_{\mathrm{anom}}(c_T,E_T^{\mathrm{min}})
\left[1-\min\!\left(\frac{|E_T|-|E_T^{\mathrm{min}}|}{B_{\mathrm{red}}},1\right)\right],\\
r_{\mathrm{drop}}(\tau) &=& \big[s_{\mathrm{anom}}(c_T,E_T)
- s_{\mathrm{anom}}(c_T,E_T\setminus E_T^{\mathrm{trig}})\big]_+,
\quad [\cdot]_+=\max(\cdot,0).
\end{array}
\end{equation}
\vspace{-0.5em}

Here, \(s_{\mathrm{anom}}(c,E)\in[0,1]\) is the anomaly confidence evaluated on the pair \((c,E)\). The threshold
\(\theta\) is the verifier's sufficiency cutoff on
\(s_{\mathrm{anom}}\), used by \(r_{\mathrm{faith}}\) as the
binarization cutoff. \(E_T^{\mathrm{min}}\) is
the smallest subset of \(E_T\) still satisfying
\(s_{\mathrm{anom}}(c_T,E')\ge\theta\), obtained by a single-pass
greedy ablation
(Appendix~\ref{appendix-a.-anomaly-efcv-diagnostics-and-verification-details}).
\(B_{\mathrm{red}}\) is the redundancy-gap budget that controls how
strictly the gap between \(E_T\) and its minimal core is penalized.
The trigger subset \(E_T^{\mathrm{trig}}\subseteq E_T\) is the subset
of windows in \(E_T\) that the policy marked as trigger evidence.

\textbf{Evidence faithfulness} \(r_{\mathrm{faith}}\) evaluates the final
decision under an evidence-only support criterion: it asks whether the
committed evidence \(E_T\) alone supports the final anomaly claim
\(c_T\). Operationally, \(r_{\mathrm{faith}}\) equals \(1\)
when \(s_{\mathrm{anom}}(c_T,E_T)\) clears the threshold \(\theta\) at
which the verifier deems the committed evidence sufficient, and \(0\)
otherwise. Binarizing through \(\theta\) keeps the faithfulness reward
stable under verifier scale drift across training.
The evidence-faithful branch in Figure~\ref{fig:method-overview}
captures this requirement: the model should justify the prediction from
\(E_T\) alone, rather than crediting an answer recoverable only from
the rollout transcript, priors, or context. This
encourages the selected evidence to serve as a
verifiable sufficient basis for the answer, not
merely as a trace left by a successful search context.

\textbf{Minimal-subset compactness} \(r_{\mathrm{min}}\)
penalizes the policy for collecting windows beyond a
sufficient core. Concretely, \(r_{\mathrm{min}}\) multiplies
\(s_{\mathrm{anom}}(c_T,E_T^{\mathrm{min}})\) by a compactness factor
\([1-\min((|E_T|-|E_T^{\mathrm{min}}|)/B_{\mathrm{red}},1)]\) that
shrinks to \(0\) as the gap \(|E_T|-|E_T^{\mathrm{min}}|\) grows
beyond the redundancy budget \(B_{\mathrm{red}}\). As the
minimal-subset branch in Figure~\ref{fig:method-overview} illustrates,
when a strict sub-core already supports the explanation, the surplus
windows are treated as wasted search budget rather than stronger
support. This trains the policy to stop searching at sufficiency
rather than exhausting the search budget.

\textbf{Drop-trigger necessity} \(r_{\mathrm{drop}}\) tests whether
trigger evidence is necessary for the final anomaly claim, rather than
merely co-selected with it. The drop-trigger branch
implements this counterfactual test: removing
\(E_T^{\mathrm{trig}}\) from \(E_T\) should reduce anomaly confidence
when the reasoning truly depends on the trigger. We target
the trigger because it is the event-chain transition at which surrounding
context becomes anomalous, whereas precursor and aftermath evidence may
be absent, redundant, or corroborative without being individually
necessary.

Together, these rewards couple sufficiency with necessity by requiring
the selected evidence to support the final claim, remain compact, and
lose support when the decisive trigger evidence is removed.

\hypertarget{seek-bench}{%
\subsection{SEEK-Bench}\label{seek-bench}}

\textbf{Event-chain design.}
An anomalous event is a temporally extended causal process: precursors
accumulate, a trigger marks the transition, and the aftermath records
the consequence. Recovering stages outside the preselected visual
bundle requires interactive acquisition, which in turn requires
stage-localizing annotations. Reasoning-enhanced VAU datasets annotate
the answer rather than the event: HIVAU-70k \citep{zhang2025holmes}
organizes instructions along temporal and semantic granularities,
whereas VAU-Bench \citep{zhu2025vau} employs a seed-reasoning
structure to standardize answer explanations. VAGU \citep{gao2026vagu}
grounds each anomaly with a temporal interval alongside its category
and QA. None of them labels the three stages, so stage-wise search has
neither a supervision target nor an evaluation contract. SEEK-Bench
annotates the event itself, providing both.

\begin{center}
\captionsetup{hypcap=false}
\includegraphics[width=\textwidth]{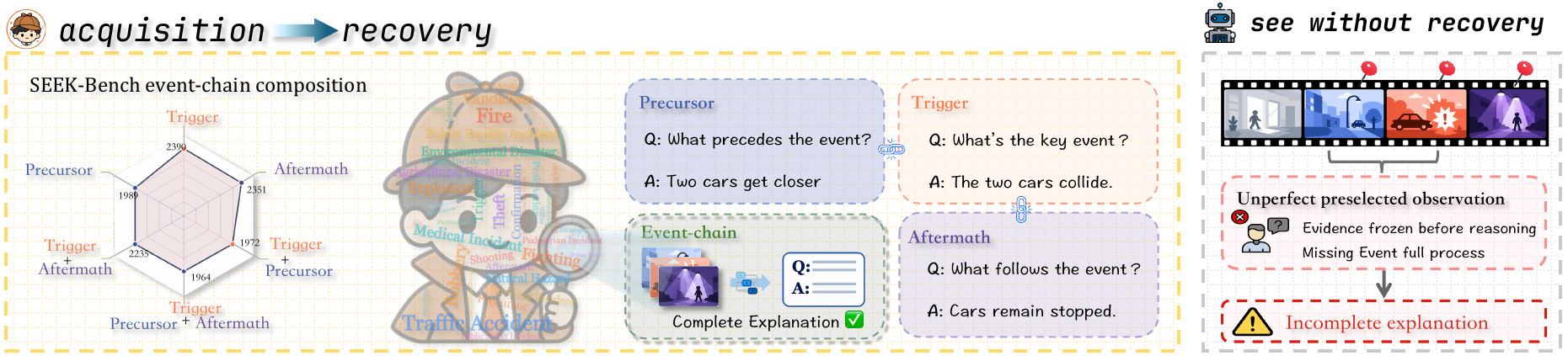}
\captionof{figure}{SEEK-Bench provides an agentic VAU evaluation contract that links anomaly episodes to temporal intervals, semantic QA and event-chain stages}
\label{fig:benchmark}
\end{center}

\textbf{Construction.} SEEK-Bench is
built from MSAD \citep{zhu2024advancing} and ECVA
\citep{du2024exploring}, whose original releases provide raw videos
with only frame-level anomaly tags. We reorganize these sources under a
unified agentic VAU contract, so that each episode supplies the five
fields a search-and-verify policy needs to consume and produce: anomaly labels, temporal intervals, semantic QA,
event-chain stage labels, and evidence windows. All
SEEK-Bench fields are annotated by human annotators.

\textbf{Statistics.} The resulting benchmark covers 21
anomaly categories plus a normal class. Figure~\ref{fig:benchmark} summarizes the data along three
axes. Figure~\ref{fig:benchmark}(a) shows the distribution over
anomaly categories. Figure~\ref{fig:benchmark}(b) visualizes
event-chain composition along two non-exclusive axes: single-stage
labels mark which of precursor, trigger, or aftermath evidence is
present, and combined labels mark episodes where the trigger co-occurs
with surrounding support. Figure~\ref{fig:benchmark}(c) shows
stage-wise duration distributions, separating short visual cues from
longer supporting windows.
Appendix~\ref{appendix-d.-seek-bench-construction-details}
provides the full construction details, including source-dataset licensing,
the manual annotation pipeline, the annotation guideline, the per-category
distribution.

\textbf{Evaluation metrics.} SEEK-Bench evaluates
search-and-verify behavior along three complementary axes: final
interval grounding, semantic case understanding, and evidence-window
recovery.
We report \textbf{Temporal mIoU} for anomaly interval
grounding, \textbf{QA Accuracy} for semantic
understanding, and \textbf{Evidence F1} for
one-to-one temporal-overlap matching between selected and reference
evidence windows.
All methods, including baselines, emit the same
structured case JSON; predicted evidence windows are extracted by one
shared deterministic parser, and episodes with no parseable citation
score zero Evidence F1. QA Accuracy uses a
semantic-equivalence judge.

\hypertarget{experiments}{%
\section{Experiments}\label{experiments}}

\hypertarget{experimental-setup}{%
\subsection{Experimental setup}\label{experimental-setup}}

\textbf{Implementation details.} We instantiate the policy with
Qwen3-VL-8B and train it in two stages. SFT uses
compact protocol traces that demonstrate search, verification, and
case finalization, while RL optimizes anomaly
rollouts with the counterfactual evidence-faithfulness rewards rather than end-label reward alone.
For RL, we use GRPO \citep{shao2024deepseekmath}, with KL coefficient \(0.1\).
The composite reward uses \(w_{\mathrm{acc}}=0.5\) and
\(w_{\mathrm{efcv}}=0.5\); inside \(R_{\mathrm{efcv}}\), we set
\(\theta=0.5\), and \(B_{\mathrm{red}}=2\). Each
episode is capped at 8 turns covering one \texttt{scan} that
uniformly samples 6 frames, three \texttt{seek}--\texttt{verify}
loops where each \texttt{seek} returns 4 frames, and one \texttt{finalize}, for a total of at most 18 sampled
frames per episode. Training runs on 8 H200 GPUs.

\textbf{Search and context budget.}
Fixed-observation baselines receive 18 uniformly sampled
full-video frames once before decoding.
SEEK-VAU instead spends this budget dynamically: the
policy begins with a protected 6-frame global scan, issues scan/seek
actions during the rollout, and keeps at most 16 visual frames visible to
the model at any decision step.
The protected scan preserves a stable full-video temporal
map for interpreting later seek evidence, while overflow removes the
oldest non-protected seek frames so adaptive search does not become
unbounded visual memory.
Internal ablations keep this active-context cap and the
turn budget fixed, changing only the training signal.

\vspace{-0.5em}
\begin{center}
\captionsetup{hypcap=false}
\captionof{table}{Comparison of Temporal mIoU and QA Accuracy across model families.}
\label{tab:main_results}
\begingroup
\setlength{\tabcolsep}{4.5pt}
\renewcommand{\arraystretch}{1.04}
\newcommand{\resultgroup}[1]{%
\specialrule{\lightrulewidth}{0pt}{0pt}
\rowcolor{gray!15}[0pt][0pt]\multicolumn{4}{@{}c@{}}{\textit{#1}}\\
}
\newcommand{\firstresultgroup}[1]{%
\rowcolor{gray!15}[0pt][0pt]\multicolumn{4}{@{}c@{}}{\textit{#1}}\\
}
\begin{tabular*}{\textwidth}{@{\extracolsep{\fill}}llcc@{}}
\toprule
\multicolumn{1}{@{}l}{\textbf{Method}} &
\multicolumn{1}{l}{\textbf{Base Model}} &
\multicolumn{1}{c}{\textbf{Temporal mIoU}} &
\multicolumn{1}{c@{}}{\textbf{QA Accuracy}} \\
\specialrule{\lightrulewidth}{0pt}{0pt}
\firstresultgroup{Video MLLMs}
GLM-4.1v \citep{hong2025glm} & GLM-4.1v-8B & {4.6} & 16.4 \\
Kimi-VL \citep{team2025kimi} & Kimi-VL-A3B & {7.3} & {16.3} \\
Intern-VL3.5 \citep{wang2025internvl3} & InternVL3.5-8B & {25.5} & {19.4} \\
Qwen3-VL \citep{bai2025qwen3} & Qwen3-VL-8B & {44.2} & 37.7 \\
\resultgroup{Video Reasoning MLLMs}
Video-R1 \citep{feng2025video} & Qwen2.5-VL-7B & {9.8} & 8.4 \\
VideoChat-R1 \citep{li2025videochat} & Qwen2.5-VL-7B & {10.0} & 13.3 \\
mimo-VL \citep{li2025xiaomi} & mimo-VL-8B & {39.4} & 16.5 \\
Ovis2.5 \citep{lu2025ovis2} & Qwen3-8B & {41.8} & 24.7 \\
\resultgroup{MLLM-based VAU Methods}
HAWK \citep{tang2024hawk} & LLaMA-2 7B & {3.7} & 19.2 \\
Holmes-VAU  \citep{zhang2025holmes} & InternVL2-2B & {7.4} & 20.7 \\
VERA \citep{ye2025vera} & InternVL2-8B & {31.5} & 18.6 \\
\resultgroup{Ours}
\textbf{SEEK-VAU (ours)} & Qwen3-VL-8B & \textbf{55.6} &
\textbf{45.5} \\
\bottomrule
\end{tabular*}
\endgroup
\end{center}
\vspace{-0.5em}

\hypertarget{main-results}{%
\subsection{Main results}\label{main-results}}

Under the SEEK-Bench protocol, Table~\ref{tab:main_results} compares
SEEK-VAU with video MLLMs and video reasoning MLLMs (18 uniformly sampled frames), and MLLM-based VAU
 (follow each method's
predefined sampling strategy) on final interval grounding and semantic case QA.

SEEK-VAU leads Table~\ref{tab:main_results} on both reported axes,
reaching 55.6 Temporal mIoU and 45.5 QA Accuracy. Relative to the
fixed-observation Qwen3-VL baseline, which achieves 44.2 and 37.7,
SEEK-VAU improves Temporal mIoU by \(+11.4\) points and QA Accuracy by
\(+7.8\) points. The gain is not confined to this baseline: Ovis2.5 is
the strongest video-reasoning baseline in Temporal mIoU and VERA is the
strongest MLLM-based VAU baseline, yet both remain below SEEK-VAU.

\vspace{-0.5em}
\begin{center}
\captionsetup{hypcap=false}
\includegraphics[width=0.8\textwidth]{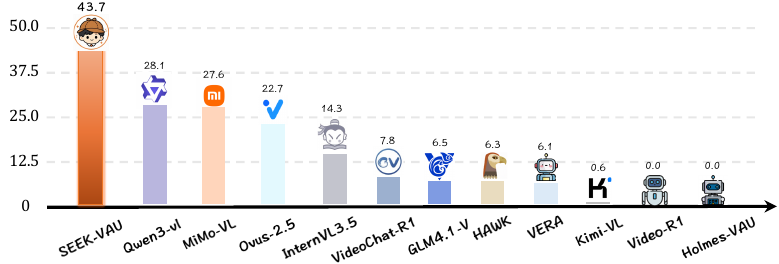}
\captionof{figure}{Comparison of Evidence F1 across models.}
\label{fig:leaderboard}
\end{center}
\vspace{-0.5em}

Figure~\ref{fig:leaderboard} shows the same advantage on Evidence F1:
SEEK-VAU reaches 43.7, ahead of the strongest non-agentic baselines
Qwen3-VL (28.1) and MiMo-VL (27.6). The advantage extends beyond final
answers and temporal intervals to the recovery of evidence windows that
justify each case. Fixed-observation baselines often predict the case
from sparse frames but miss those windows. SEEK-VAU narrows this gap by
issuing later \texttt{seek} calls to fill missing evidence and
verifying it before \texttt{finalize}, while keeping the same
active-context budget. Crucially, SEEK-VAU rarely uses the full search
budget: the average frame usage stays well below the 18-frame cap,
whereas fixed-observation baselines always consume their full
allocation. The search-and-verify policy is therefore more accurate
while using fewer visual frames, with modest token and latency
overhead.

\hypertarget{ablation-studies-and-analysis}{%
\subsection{Ablation studies and
analysis}\label{ablation-studies-and-analysis}}

We report an attribution ablation
(Table~\ref{tab:attribution}), an acquisition-mode ablation
(Table~\ref{tab:acquisition}), a visual-context cap ablation
(Table~\ref{tab:visual_budget_ablation}).

\noindent
\captionsetup{hypcap=false}
\begin{minipage}[t]{0.58\textwidth}
\centering
\captionof{table}{Attribution ablation.}
\label{tab:attribution}
\begingroup
\renewcommand{\arraystretch}{1.05}
\setlength{\tabcolsep}{2.5pt}
\small
\begin{tabular}{@{}lccc@{}}
\toprule
\textbf{Method} & \textbf{T. mIoU} & \textbf{QA Acc.} &
\textbf{E.F1} \\
\midrule
Fixed obs.\ + basic sup. & 44.2 & 37.7 & 28.1 \\
Fixed obs.\ + rich sup. & 46.9 & 38.2 & 28.3 \\
Active + basic sup. & 50.1 & 40.1 & 34.3 \\
Active + rich sup.\ (w/o EFCV) & 54.4 & 42.7 & 41.6 \\
\textbf{SEEK-VAU (full)} & \textbf{55.6} &
\textbf{45.5} & \textbf{43.7} \\
\bottomrule
\end{tabular}
\endgroup
\end{minipage}
\hfill
\begin{minipage}[t]{0.40\textwidth}
\centering
\captionof{table}{Visual-context cap ablation.}
\label{tab:visual_budget_ablation}
\begingroup
\renewcommand{\arraystretch}{1.05}
\small
\begin{tabular}{@{}cccc@{}}
\toprule
\textbf{Cap} & \textbf{T. mIoU} & \textbf{QA Acc.} &
\textbf{E.F1} \\
\midrule
12 & 51.6 & 44.1 & 38.2 \\
14 & {53.2} & {43.8} & {39.3} \\
16 & \textbf{{55.6}} & {45.5} & \textbf{{43.7}} \\
18 & {53.6} & {45.7} & {38.2} \\
20 & 55.3 & \textbf{{45.9}} & {42.8} \\
\bottomrule
\end{tabular}
\endgroup
\end{minipage}
\par\medskip

\textbf{Attribution ablation.} Do the gains stem from the
search-and-verify protocol or from the richer supervision that accompanies
SEEK-Bench? Table~\ref{tab:attribution} disentangles the two sources under
a matched backbone and 18-frame visual budget. Basic
supervision trains on task fields only (existence, closed-ended QA, etc.), while rich supervision adds event-chain stage labels and
evidence windows.
Supervision alone explains little: under fixed observation, upgrading
from basic to full event-chain annotations lifts T.\ mIoU
by only 2.7 points and moves QA Acc.\ and E.F1 by a negligible 0.5 and
0.2, teaching the model
\textbf{\emph{what}} to report rather than \textbf{\emph{where}} to look. The acquisition
protocol is the stronger single factor (replacing fixed
observation with active search, again without event-chain supervision,
raises the three metrics by 5.9, 2.4, and 6.2 points respectively);
combined with event-chain supervision the two are super-additive (41.6
E.F1, 7.1 points beyond the sum of the individual gains,
and active search under matched event-chain supervision improves T.\ mIoU,
QA Acc., and E.F1 by 7.5, 4.5, and 13.3 points over the fixed
rich-supervision control), and EFCV yields $+2.8$ QA
Acc. Acquisition thus drives evidence recovery, while supervision improves QA Acc by $+2.6$ under active acquisition, amplifying rather
than replacing the search-and-verify behavior.

\textbf{Visual-context cap ablation.}
Within the examined range, raising the cap from 12 to 16 improves all
three metrics (T.\,mIoU 51.6 to 55.6, QA Acc.\ 44.1 to 45.5, and E.F1
38.2 to 43.7; Table~\ref{tab:visual_budget_ablation}), with the
largest gain in Evidence F1, while QA Acc.\ dips only slightly between
caps 12 and 14 (44.1 to 43.8). Beyond
16 frames the trend is non-monotone: cap 18 gains marginal QA Accuracy
(45.7) but drops T.\,mIoU to 53.6 and E.F1 back to the cap-12 level
(38.2), and cap 20 partially recovers (55.3/45.9/42.8), attaining the
best QA Accuracy yet remaining below the 16-frame peak on the other two
metrics. The protocol's 16-frame cap thus best balances evidence
recovery and compact visual context.

\begin{center}
\captionsetup{hypcap=false}
\captionof{table}{Matched acquisition-mode ablation.}
\label{tab:acquisition}
\begingroup
\renewcommand{\arraystretch}{1.05}
\setlength{\tabcolsep}{6pt}
\small
\begin{tabular}{@{}lcccccc@{}}
\toprule
\textbf{Method} & \textbf{Acquisition} & \textbf{Verify} & \textbf{T. mIoU} &
\textbf{QA Acc.} & \textbf{E.F1} & \textbf{Frames} \\
\midrule
Two-stage matched & One-shot & $\times$ & 48.2 & 31.1 & 32.3 & 18.0 \\
Closed-loop w/o verify & Closed-loop & $\times$ &
50.7 & 33.5 & 29.1 & 13.2 \\
SEEK-VAU w/o EFCV & Closed-loop & $\surd$ &
54.4 & 42.7 & 41.6 & 17.1 \\
\textbf{SEEK-VAU (full)} & Closed-loop & $\surd$ &
\textbf{55.6} & \textbf{45.5} & \textbf{43.7} & 15.6 \\
\bottomrule
\end{tabular}
\endgroup
\end{center}

\textbf{Acquisition-mode ablation.} Does the gain come
from the closed-loop search-and-verify protocol itself, or from any
form of adaptive acquisition? Table~\ref{tab:acquisition} compares
matched protocols in which only the acquisition protocol, the verify
gate, and the EFCV reward differ, while the backbone, training
episodes and targets, update count, answer reward, 18-frame cap,
decoding, parser, and evaluator are held fixed; \emph{Frames} reports
realized average use rather than the shared cap. Against the
fixed-observation \emph{Fixed obs.\ + rich sup.} rung of
Table~\ref{tab:attribution}, one-shot adaptive retrieval improves
T.\ mIoU by only $+1.3$ while QA Acc.\ drops by $7.1$: frozen
retrieval alone cannot explain the gain. Closing the loop without
verification lifts T.\ mIoU by a further $+2.5$ yet \emph{lowers}
E.F1 by $3.2$ while spending only 13.2 frames: the policy re-plans
well but closes the case on an incomplete evidence ledger. Gating
finalization on verification is the decisive step: it adds 3.7 points
of T.\ mIoU, 9.2 of QA Acc., and 12.5 of E.F1 while raising realized
use to 17.1 frames, exactly the behavior expected of a verifier that
withholds finalization until the evidence is sufficient rather than
re-scoring a finished answer. EFCV adds a further 1.2, 2.8, and 2.1
points on the same three metrics while \emph{reducing} average use to
15.6 frames, so the full policy is both more accurate and more
frugal.

\vspace{-0.5em}
\begin{center}
\captionsetup{hypcap=false}
\captionof{table}{Inference efficiency ablation.}
\label{tab:efficiency}
\begingroup
\renewcommand{\arraystretch}{1.05}
\small
\begin{tabular}{@{}lcccc@{}}
\toprule
\textbf{Model} & \textbf{Avg. frames} & \textbf{Avg. tokens} &
\textbf{Latency (s)} & \textbf{Throughput (ep/min)} \\
\midrule
Qwen3-VL         & 18.0 & 4.33k & 8.71 & 20.22 \\
SEEK-VAU w/o EFCV     & 17.1 & 7.17k & 10.32 & 18.45 \\
SEEK-VAU      & 15.6 & 5.78k & 9.27 & 19.31 \\
\bottomrule
\end{tabular}
\endgroup
\end{center}
\vspace{-0.3em}

\textbf{Inference-time efficiency.} A central claim of
SEEK-VAU is that an agentic search-and-verify protocol can operate
under a bounded search and context budget.
Table~\ref{tab:efficiency} corroborates this: SEEK-VAU consumes only
15.6 frames on average, well below the 18-frame ceiling and roughly
13\% fewer than the fixed-observation Qwen3-VL baseline (18.0 frames,
4.33k tokens, 8.71\,s, 20.22\,ep/min). The budget alone is not
sufficient: without EFCV, the policy nearly saturates it at 17.1 frames and
inflates token usage to 7.17k with the worst latency of 10.32\,s. By
rewarding selective commitment to key evidence rather than exhaustive
coverage, EFCV cuts token usage relative to the w/o-EFCV variant by
roughly 19\% to 5.78k, yielding 9.27\,s latency and 19.31\,ep/min.
Together, the bounded search budget and EFCV keep multi-turn
deliberation practical under a fixed visual-context cap.
\vspace{-0.5em}

\hypertarget{case-study}{%
\subsection{Case study}\label{case-study}}

\vspace{-0.3em}
\begin{center}
\captionsetup{hypcap=false}
\includegraphics[width=\textwidth]{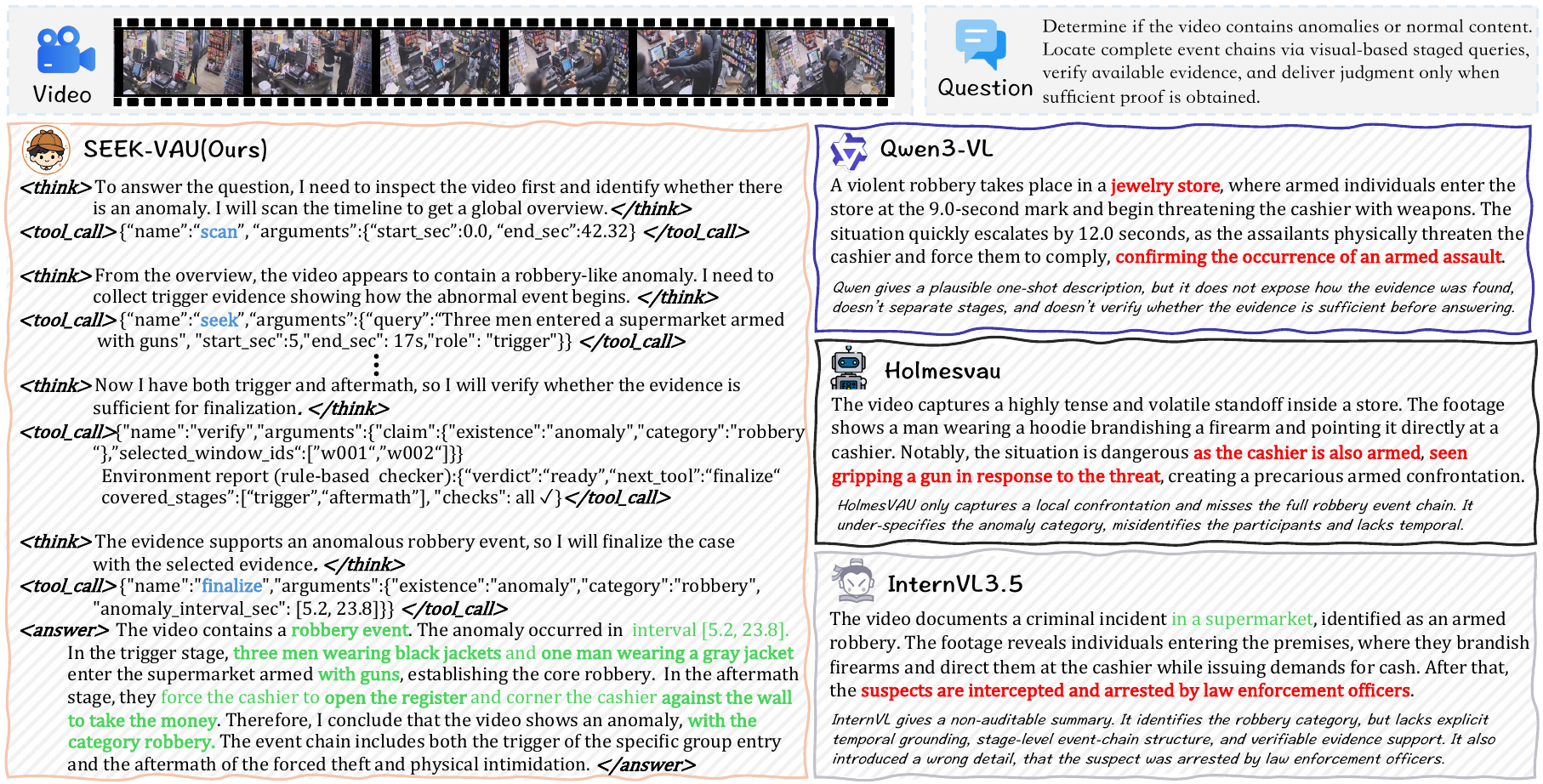}
\captionof{figure}{Case study of SEEK-VAU on a supermarket robbery
video.}
\label{fig:case-study}
\end{center}
\vspace{-0.3em}

Figure~\ref{fig:case-study} compares free-text anomaly description
with evidence-grounded case construction on a supermarket robbery.
All fixed-observation baselines label the event robbery but
diverge in detail: Qwen3-VL misattributes the scene to a jewelry
store and narrows to a 9--12\,s confrontation; Holmes-VAU
hallucinates the cashier as also armed and reduces the case to a
local standoff; InternVL3.5 fabricates an arrest by law enforcement
absent from the video. SEEK-VAU instead produces a traceable
trajectory: a global scan, stage-tagged \texttt{seek} calls
for trigger and aftermath, and \texttt{finalize} only after
verification confirms both stages. The resulting case grounds the
robbery to \([5.2, 23.8]\,\)s, with group entry as trigger and
forced register opening with physical intimidation as aftermath.
The episode has no visible precursor, and SEEK-VAU correctly does
not fabricate one. Both the temporal interval and the stage
attribution trace back to specific \texttt{seek} rather than
to free-text generation.
\vspace{-0.5em}

\hypertarget{conclusion}{%
\section{Conclusion}\label{conclusion}}

\vspace{-0.5em}
We present SEEK-VAU, an agentic search-and-verify protocol under a
bounded search budget for evidence-faithful video anomaly
understanding. Rather than decoding from a fixed observation set,
SEEK-VAU gathers evidence, verifies its readiness, and finalizes an
event chain over precursor, trigger, and aftermath. The budget caps
per-turn input and cumulative context, while EFCV trains selective
key-evidence search, so backtracking improves recovery with 
modest inference overhead. With SEEK-Bench, evidence acquisition,
verification, and bounded-cost search become explicit training and
evaluation targets, advancing VAU toward verifiable and efficient anomaly reasoning.

\section*{Acknowledgments and Disclosures}
\textbf{Funding.}
This work was supported by the National Natural Science Foundation of China under Grants U22B2053 and 62573393.

\textbf{Competing interests.} The authors have no competing interests to disclose.

\bibliographystyle{plainnat}
\bibliography{ref}

\newpage
\appendix
\section{Evidence F1 computation}
\label{app:evidence-f1}

Evidence windows are committed by \texttt{seek} actions: each
\texttt{seek} return commits exactly one window, the interval from its
earliest to its latest returned frame
(Section~\ref{interleaved-event-chain-search}), and distinct
\texttt{seek} calls commit distinct windows even when their intervals
overlap. The three \texttt{seek} calls available within the episode
budget therefore commit at most three windows, which is the evidence
budget \(|E_T|\le 3\) used by the EFCV reward
(Appendix~\ref{appendix-a.-anomaly-efcv-diagnostics-and-verification-details}).

For the default predicted evidence set
\(\hat E_{\mathrm{top}}=\{\hat t_i\}_{i=1}^{\min(3,|\hat E|)}\) and annotated
evidence windows \(E^\star=\{t_j\}\), we define

\begin{equation}
M=\max_{\pi\in\Pi}
\sum_{(i,j)\in\pi}
\mathbf{1}\!\left[\mathrm{IoU}(\hat t_i,t_j)\ge\tau_t\right].
\end{equation}
\begin{equation}
P=\frac{M}{|\hat E_{\mathrm{top}}|},\qquad
R=\frac{M}{|E^\star|},\qquad
\mathrm{Evidence\ F1}=\frac{2PR}{P+R}.
\end{equation}

Here, \(\Pi\) denotes one-to-one matchings between predicted and
annotated evidence windows, and samples without valid predicted or
annotated evidence receive zero Evidence F1.
We set $\tau_t=0.3$ throughout, and the annotated set
$E^\star$ comprises each stage's primary reference window plus any
adjudicated admissible alternatives
(Appendix~\ref{appendix-d.-seek-bench-construction-details}). Evidence
F1 therefore measures one-to-one recovery of adjudicated canonical
reference windows, which are canonical representative windows rather
than an exhaustive set of valid alternatives; it does not by itself
verify actor or object identity or complete semantic faithfulness.

\hypertarget{appendix-a.-anomaly-efcv-diagnostics-and-verification-details}{%
\section{Anomaly EFCV Diagnostics and Verification Details}\label{appendix-a.-anomaly-efcv-diagnostics-and-verification-details}}

This appendix specifies how the anomaly-support scalar
\(s_{\mathrm{anom}}\) used in the EFCV reward
(Section~\ref{efcv-learning}) is obtained from the verify action
contract of Section~\ref{policy-internal-evidence-verification}, and
maps the EFCV reward terms to their implementation labels.

Each verify call emits a scalar
\(s_{\mathrm{anom}}\in[0,1]\); we write \(s_{\mathrm{anom}}(c,E)\) for
the value emitted when the verify contract is invoked on the pair
\((c,E)\). Claim sufficiency, decision-critical necessity, and closure
readiness are consolidated into this one scalar for reward computation;
the additional diagnostic fields emitted by the verify schema
(Appendix~\ref{verifier-prompts}) do not enter the reward. Evaluating
\(s_{\mathrm{anom}}\)
on the perturbed sets \(E_T^{\mathrm{min}}\) and
\(E_T \setminus E_T^{\mathrm{trig}}\) reduces to re-invoking the same
contract on the perturbed inputs. \(s_{\mathrm{anom}}\) remains a
diagnostic support scalar, not a formal proof of sufficiency.
Together, the three reward terms train the policy to spend the
bounded search budget selectively on key evidence rather than
exhaustively on full coverage: \(r_{\mathrm{faith}}\) requires the
committed evidence set as a whole to support closure, \(r_{\mathrm{min}}\) drives the
policy to stop at a minimal sufficient subset, and \(r_{\mathrm{drop}}\)
rewards trajectories whose support weakens when
trigger evidence is removed.

For anomaly rollouts, the EFCV reward \(R_{\mathrm{efcv}}\) of
Section~\ref{efcv-learning} instantiates as

\begin{equation}
\begin{gathered}
R_{\mathrm{efcv}}(\tau)
= r_{\mathrm{faith}}(\tau)+r_{\mathrm{min}}(\tau)+r_{\mathrm{drop}}(\tau),\\
\left\{
\begin{array}{@{}l@{\;}c@{\;}l@{}}
r_{\mathrm{faith}}(\tau) &=& \mathbf{1}\!\left[s_{\mathrm{anom}}(c_T,E_T)\ge\theta\right],\\
r_{\mathrm{min}}(\tau) &=& s_{\mathrm{anom}}(c_T,E_T^{\mathrm{min}})\left[1-\min\!\left(\frac{|E_T|-|E_T^{\mathrm{min}}|}{B_{\mathrm{red}}},1\right)\right],\\
r_{\mathrm{drop}}(\tau) &=& \big[s_{\mathrm{anom}}(c_T,E_T)-s_{\mathrm{anom}}(c_T,E_T\setminus E_T^{\mathrm{trig}})\big]_+,
\end{array}
\right.
\end{gathered}
\end{equation}

The three terms correspond to evidence faithfulness, minimal-subset
compactness, and drop-trigger necessity. Here \(\theta\)
is the verifier-deems-sufficient threshold shared across the EFCV
terms: \(r_{\mathrm{faith}}\) is the indicator that the committed
\(E_T\) clears \(\theta\), and
\(E_T^{\mathrm{min}}=\arg\min_{E'\subseteq E_T,\, s_{\mathrm{anom}}(c_T,E')\ge\theta}|E'|\) is the smallest
subset still satisfying \(s_{\mathrm{anom}}(c_T,E')\ge\theta\); \(B_{\mathrm{red}}>0\) is the
tolerated redundancy-gap budget, \(E_T^{\mathrm{trig}}\) is the subset
of selected evidence assigned to the trigger stage, and
\([x]_+=\max(x,0)\). If no subset satisfies the minimal-subset
constraint, \(r_{\mathrm{min}}\) is set to \(0\) by convention. The
minimal-subset term penalizes selected evidence that is much larger than
its compact sufficient core. The drop-trigger term measures whether
removing trigger-stage evidence weakens anomaly support. The
evidence-faithfulness term scores whether the committed selected
evidence alone supports the final claim after rollout completion.

The table below summarizes the implementation labels corresponding to
the paper-level EFCV terms.

\begin{longtable}[]{@{}
  >{\raggedright\arraybackslash}p{(\columnwidth - 4\tabcolsep) * \real{0.33}}
  >{\raggedright\arraybackslash}p{(\columnwidth - 4\tabcolsep) * \real{0.33}}
  >{\raggedright\arraybackslash}p{(\columnwidth - 4\tabcolsep) * \real{0.33}}@{}}
\toprule
Paper term & Implementation label & Role \\
\midrule
\endhead
\(r_{\mathrm{faith}}\) & evidence faithfulness & Scores the committed
claim-evidence pair using committed selected evidence only \\
\(r_{\mathrm{min}}\) & \texttt{minimal\_subset} & Tests whether a
compact sufficient core exists without excessive redundant selected
evidence \\
\(r_{\mathrm{drop}}\) & \texttt{drop\_trigger} & Tests whether
trigger-stage evidence is decision-critical \\
\bottomrule
\end{longtable}

\paragraph{Hyperparameter values used in our experiments.}
These values are held fixed across all reported runs.

\begin{longtable}[]{@{}
  >{\raggedright\arraybackslash}p{(\columnwidth - 4\tabcolsep) * \real{0.22}}
  >{\raggedright\arraybackslash}p{(\columnwidth - 4\tabcolsep) * \real{0.18}}
  >{\raggedright\arraybackslash}p{(\columnwidth - 4\tabcolsep) * \real{0.50}}@{}}
\toprule
Symbol & Value & Role \\
\midrule
\endhead
\(w_{\mathrm{acc}}\) & 0.5 & Weight on the answer-correctness reward
\(R_{\mathrm{acc}}\) \\
\(w_{\mathrm{efcv}}\) & 0.5 & Weight on the evidence-faithfulness reward
\(R_{\mathrm{efcv}}\) \\
\(\theta\) & 0.5 & Verifier-deems-sufficient threshold; gates \(r_{\mathrm{faith}}\) (via the indicator) and the compact-sufficient core \(E_T^{\mathrm{min}}\) \\
\(B_{\mathrm{red}}\) & 2 & Redundancy budget allowing up to two extra evidence windows beyond the minimal sufficient core \\
\bottomrule
\end{longtable}

\paragraph{\rev{Answer-correctness reward $R_{\mathrm{acc}}$.}}
\rev{$R_{\mathrm{acc}}$ aggregates three task-level scores by
unweighted sum:
\begin{equation}
R_{\mathrm{acc}}(\tau)=s_{\mathrm{cat}}(\tau)+s_{\mathrm{tIoU}}(\tau)+s_{\mathrm{qa}}(\tau),
\end{equation}
where $s_{\mathrm{cat}}(\tau)\in\{0,1\}$ is the indicator that the finalized
anomaly category matches the ground-truth label, $s_{\mathrm{tIoU}}(\tau)\in[0,1]$
is the temporal IoU between the finalized interval and the annotated anomaly
interval, and $s_{\mathrm{qa}}(\tau)\in[0,1]$ is the per-episode average of
binary semantic-equivalence scores returned by the QA judge over the
episode's QA pairs. We use the same GPT-5.4 judge model and prompt as at
evaluation (Appendix~\ref{appendix-g.-prompts}, QA semantic-equivalence
judge), so that $R_{\mathrm{acc}}$ during RL and QA Accuracy at evaluation
share identical scoring. Each summand lies in $[0,1]$, placing
$R_{\mathrm{acc}}\in[0,3]$ on the same range as
$R_{\mathrm{efcv}}=r_{\mathrm{faith}}+r_{\mathrm{min}}+r_{\mathrm{drop}}\in[0,3]$,
which is what makes the equal weighting
$w_{\mathrm{acc}}=w_{\mathrm{efcv}}=0.5$ comparable across rewards.}

\paragraph{Computing \(E_T^{\mathrm{min}}\).}
We use a single-pass greedy algorithm: starting from the full selected
evidence set \(E_T\), we rank windows by verifier-induced marginal
contribution and remove them from lowest to highest contribution while
recomputing the support score after each removal. The last subset \(E'\)
that still satisfies \(s_{\mathrm{anom}}(c_T,E') \ge \theta\) is taken
as \(E_T^{\mathrm{min}}\). This requires
\(O(|E_T|^2)\) verifier queries per rollout and is tractable in our
setting because \(|E_T| \le 3\) (bounded by the three \texttt{seek}
calls available within the eight-turn episode budget). For
larger evidence sets, one could instead use an LP relaxation or a
submodular-style greedy approximation with the usual \(1-1/e\)
guarantee under the corresponding surrogate assumptions, but that
regime is unnecessary here.

\paragraph{\rev{Qualitative monotonicity argument for redundancy control.}}
\rev{We do not claim a closed-form bound on expected
over-collection; instead, $r_{\mathrm{min}}$ controls redundancy by
qualitative monotonicity. Let $\rho(E)=|E|-|E_T^{\mathrm{min}}|$ denote the
redundancy gap of a selected evidence set relative to its greedy
support-preserving core. By construction of $r_{\mathrm{min}}$, the
compactness factor $[1-\min(\rho(E)/B_{\mathrm{red}},1)]$ is non-increasing
in $\rho(E)$ and saturates to zero once $\rho(E)>B_{\mathrm{red}}$. Any
policy that consistently selects an evidence set with
$\rho(E)>B_{\mathrm{red}}$ can therefore match or improve its
$r_{\mathrm{min}}$ score by discarding windows down to a sufficient subset,
holding $r_{\mathrm{faith}}$ and $r_{\mathrm{drop}}$ unchanged when the
discarded windows are non-decisive. The magnitude of the resulting reward
change depends on the verifier's empirical sensitivity to evidence-set
perturbations, which we characterize through the per-component EFCV ablation
in Appendix~\ref{appendix-efcv-component-ablation} rather than through an
analytical bound.}

\hypertarget{appendix-f.-pomdp-formalization}{%
\section{\rev{Stop-Action Decision Process: Notation}}\label{appendix-f.-pomdp-formalization}}

\rev{This appendix introduces the notation used to
describe the four-action protocol of Section~\ref{task-formulation};
it is a notational scaffold for cross-referencing the protocol
components, not a theoretical contribution. We do not claim a regret
bound, sample-complexity bound, or optimal-stopping guarantee, and we
do not invoke any classical sequential-analysis result to justify the
training procedure.} We describe the protocol of Section~\ref{task-formulation} as a
stop-action decision process
\(\mathcal{M}=(\mathcal{S},\mathcal{A},\mathcal{O},T,R,\gamma,\mathcal{F})\),
where the action set
\(\mathcal{A}=\{\texttt{scan},\texttt{seek},\texttt{verify},\texttt{finalize}\}\)
interleaves evidence acquisition and verification over an observation
manifold \(\mathcal{O}\) of RGB frames and verifier outputs, the
reward \(R\) is the EFCV-grounded objective of
Section~\ref{efcv-learning}, and \texttt{finalize}\(\in\mathcal{F}\) is
the lone stopping action committed via a learned stopping policy
\(\pi_{\mathrm{stop}}:\mathcal{O}\to[0,1]\) that consults
the verifier-emitted scalar \(s_{\mathrm{anom}}\) and the recommended
action \(a_t\) before \texttt{finalize} becomes admissible. \rev{Fixed-observation VAU is recovered as the degenerate single-step
regime when \(\mathcal{A}=\{\texttt{finalize}\}\); we use this only as
a contrast for protocol description.} The observation manifold
\(\mathcal{O}\) is constrained by the bounded search budget: each
interaction step admits at most a fixed-size visual context, and a
rolling active-context window evicts older non-protected frames
before any new \texttt{seek} output is admitted, so the transition
function \(T\) enforces budget-bounded observation history rather
than unbounded accumulation.

\hypertarget{appendix-d.-seek-bench-construction-details}{%
\section{\rev{SEEK-Bench Construction Details}}\label{appendix-d.-seek-bench-construction-details}}

\hypertarget{source-dataset-licensing-pathway}{%
\subsection{\rev{Source-Dataset Licensing Pathway}}\label{source-dataset-licensing-pathway}}

\revblock{SEEK-Bench inherits its licensing pathway from MSAD and ECVA: our
release redistributes only the derived structured annotations, while the
underlying videos, split identity, and access conditions remain governed
by the source datasets. In practice, this means that benchmark users are
expected to obtain the original assets under the upstream dataset terms
and then align them with the SEEK-Bench episode metadata and evidence
annotations.}

\hypertarget{manual-annotation-pipeline}{%
\subsection{\rev{Manual Annotation}}\label{manual-annotation-pipeline}}

GPT-5.4 proposals are used only to initialize the
annotation workflow. Each canonical field is independently verified and
adjudicated by human annotators before it is eligible for a
deterministic training/evaluation release.

{
\begin{table}[h]
\centering
\small
\caption{Source and field provenance for SEEK-Bench.}
\label{tab:appendix-d-field-provenance}
\begin{tabular}{@{}p{0.23\textwidth}p{0.19\textwidth}p{0.22\textwidth}p{0.25\textwidth}@{}}
\toprule
\textbf{Field} & \textbf{Source status} & \textbf{SEEK contribution} & \textbf{Production and use} \\
\midrule
Anomaly status/category & MSAD/ECVA base labels & Canonical mapping & Source label plus human check; QA/classification \\
Temporal interval & Partial or absent upstream & Added & Human annotation; Temporal mIoU \\
Event-chain stage & Not upstream & Added & Human annotation; protocol supervision \\
Semantic QA & Not upstream & Added & Human authored and independently verified; QA Accuracy \\
Evidence window & Not upstream & Added & Human selection and audit; Evidence F1 \\
Provenance metadata & Upstream identifiers & SEEK manifest & Deterministic; split and leakage audits \\
\bottomrule
\end{tabular}
\end{table}
}

\hypertarget{annotation-guideline-key-rules}{%
\subsection{\rev{Annotation Guideline (Key Rules)}}\label{annotation-guideline-key-rules}}

\begin{itemize}
\tightlist
\item
\revblock{\textbf{Temporal interval boundary rule.} An interval starts at the first
visible cue that materially contributes to the anomaly account and ends
at the point of clear return to a normal or post-event steady state.}
\item
\revblock{\textbf{Stage-presence rule.} The trigger stage is required, while
precursor and aftermath stages are optional and are annotated only when
they are visibly supported by the episode.}
\item
\revblock{\textbf{Stage-overlap rule.} Stages are annotated as non-overlapping
intervals; when a short ambiguous region could belong to two stages, it
is assigned to the dominant stage in the adjudicated record.}
\item
\revblock{\textbf{Evidence-moment rule.} Evidence moments must be strict
sub-windows of an annotated stage and should isolate the smallest
visually sufficient moment that justifies the stage label.}
\item
\revblock{\textbf{QA-grounding rule.} Each QA pair must be answerable from the
annotated stages and evidence moments alone; questions that require
external world knowledge or off-video assumptions are rejected.}
\item
\revblock{\textbf{Category-canonicalization rule.} Free-form anomaly labels are
canonicalized into a fixed 22-label vocabulary used by the released
JSONL and evaluator. Table~\ref{tab:appendix-d-category-distribution}
lists the categories, while the release files retain the corresponding
\texttt{snake\_case} keys.}
\end{itemize}

\hypertarget{category-distribution-table}{%
\subsection{\rev{Category Distribution Table}}\label{category-distribution-table}}

\begin{table}[h]
\centering
\small
\caption{Canonical category vocabulary used by SEEK-Bench release files
and evaluator. Counts are episode-level totals across the train and
val splits of the current release candidate (2{,}878 episodes;
2{,}390 anomaly + 488 normal).}
\label{tab:appendix-d-category-distribution}
\begin{tabular}{@{}p{0.30\textwidth}cp{0.30\textwidth}c@{}}
\toprule
\textbf{Category} & \textbf{Count} & \textbf{Category} & \textbf{Count} \\
\midrule
Normal & 488 & Natural hazard & 127 \\
Agricultural disaster & 102 & Pedestrian incident & 75 \\
Animals hunting people & 7 & Protests & 23 \\
Assault & 15 & Public facility incident & 108 \\
Building incident & 88 & Public safety incident & 79 \\
Environmental disaster & 175 & Robbery & 232 \\
Explosion & 79 & Shooting & 29 \\
Fighting & 190 & Theft & 72 \\
Fire & 259 & Traffic accident & 438 \\
Food safety incident & 62 & Vandalism & 85 \\
Medical incident & 100 & Water incident & 45 \\
\bottomrule
\end{tabular}
\end{table}

\hypertarget{train-val-split}{%
\subsection{\rev{Train/Val Split and Disjointness}}\label{appendix-d-train-val-split}}

\revblock{The 2{,}878-episode release candidate is partitioned into a train/val
split with an approximate 8:2 ratio at the episode level. The split is
\emph{video-disjoint}: every source video and all of its derived episodes are
assigned to a single side of the partition, so no source video contributes
to both sides. Within this video-disjoint constraint, the split is stratified
by anomaly category so that the per-category proportions remain close to
those reported in Table~\ref{tab:appendix-d-category-distribution}.}

\revblock{The resulting episode counts are summarized in
Table~\ref{tab:appendix-d-train-val-split}: 2{,}302 train episodes
(1{,}912 anomaly + 390 normal) and 576 val episodes
(478 anomaly + 98 normal).}

\begin{table}[h]
\centering
\caption{\rev{Episode-level train/val split for the SEEK-Bench
release candidate. The split is video-disjoint and stratified by anomaly
category; ratios are reported relative to the 2{,}878-episode total.}}
\label{tab:appendix-d-train-val-split}
\begin{tabular}{lcccc}
\toprule
\textbf{Subset} & \textbf{Anomaly} & \textbf{Normal} & \textbf{Total} & \textbf{Ratio} \\
\midrule
Train & 1{,}912 & 390 & 2{,}302 & 80.0\% \\
Val   & 478   & 98  & 576   & 20.0\% \\
\midrule
All   & 2{,}390 & 488 & 2{,}878 & 100.0\% \\
\bottomrule
\end{tabular}
\end{table}

\revblock{All evaluation numbers reported in the main paper
(Tables~\ref{tab:main_results}, \ref{tab:attribution},
\ref{tab:visual_budget_ablation}, \ref{tab:acquisition},
\ref{tab:efficiency} and
Figure~\ref{fig:leaderboard}) are computed on the val split. The SFT trace
construction described in Appendix~\ref{appendix-sft-trace-construction}
draws exclusively from the train split, and the GRPO RL stage rolls out only
on train episodes, so val episodes are unseen by both training stages.}

\hypertarget{seek-bench-dataset-card}{%
\subsection{SEEK-Bench Dataset Card}
\label{seek-bench-dataset-card}}

\paragraph{Source provenance and filtering.}
SEEK-Bench is derived from MSAD and ECVA. The current
release redistributes only SEEK's derived structured annotations and
episode metadata; it does not redistribute the source videos. Users must
obtain the underlying assets under the terms specified by the upstream
datasets. The MSAD source is identified by its official repository
(\url{https://github.com/Tom-roujiang/MSAD}), which specifies
academic-only use and restrictions on commercial use, reproduction,
distribution, and sale. The ECVA source is identified by
\url{https://github.com/dulpy/ECVA}; its repository does not specify a
separate license in the checked record, so its upstream terms apply.
Exact source versions, filtering and deduplication rules, and any privacy
notes are recorded with the released provenance manifest; we do not infer
or assign an additional license to the upstream videos.

\paragraph{Release and reproducibility metadata.}
Counts in the tables above refer to the frozen release
snapshot. The canonical release manifest records source identifiers,
video-level split membership, annotation-schema and evaluator versions,
and a manifest checksum.

\paragraph{Independent annotation and adjudication.}
Model proposals initialize candidate fields only. A
seven-person pool with computer-vision or multimodal backgrounds
independently verifies and corrects every episode with two annotators per
episode. Annotators verify anomaly existence, category, temporal
interval, applicable stages, evidence windows, and QA answerability; they
do not receive policy reward values or policy-generated rationales.
Three senior VAU adjudicators resolve unresolved disagreements and record
the final canonical decision. Only records passing provenance and schema
checks enter the released training/evaluation manifest.

\paragraph{Annotation agreement.}
Agreement is computed before adjudication, on applicable
independent records only. Precursor and aftermath presence have
inter-annotator agreement $\alpha=0.92$ and $\alpha=0.89$, respectively.
For stages marked by both annotators, the median pairwise boundary tIoU
is $0.78$, and $85\%$ of pairs reach tIoU $\geq 0.5$. Semantic-QA
agreement is $\alpha=0.91$ with $93\%$ raw agreement. For evidence
windows, agreement is likewise computed before adjudication: under the
Evidence F1 operating rule (one-to-one tIoU matching at $\tau_t=0.3$),
the cross-annotator Evidence F1 between the two annotators' window sets
is $0.81$ averaged over $2{,}390$ anomalous episodes. Among window pairs
matched across annotators within the same stage role, the median tIoU is
$0.84$ for trigger, $0.79$ for precursor, and $0.81$ for aftermath
evidence ($91\%$, $85\%$, and $87\%$ of pairs reach tIoU $\geq 0.5$;
unmatched window fractions $6\%$, $9\%$, and $8\%$). The two annotators
mark the same number of windows in $90\%$ of applicable stage cells. All
agreement statistics quantify annotation stability of the released
reference; they are computed before adjudication and do not involve
model outputs.

\paragraph{QA construction and validation.}
The QA suite is stage-grounded rather than free-form:
anomaly episodes carry eight typed questions covering case-level
attributes (existence, category, temporal interval, severity),
chain-stage evidence (trigger action, precursor timing when a precursor
is annotated, earliest-warning frame), and a counterfactual check, so
episodes without an annotated precursor carry seven questions; normal
episodes carry a five-question normal-specific suite covering existence,
category, temporal interval, earliest-warning frame, and normal reason.
Anomaly-suite questions are authored per episode with entity-specific
phrasing rather than drawn from a shared template bank. Question writing
and answer verification are separate steps. A question is retained only
when independent reviewers judge it visually answerable, sufficiently
specific, and answerable from the annotated stages and evidence moments
without external knowledge. Unanswerable, ambiguous, duplicate, or
evidence-inconsistent questions are removed or adjudicated; normal
episodes use normal-specific QA. The QA agreement statistic is computed
between the two human annotators, not against model outputs. QA Accuracy
is computed by a common GPT-5.4 binary semantic judge with the same
prompt, temperature-0 decoding, and parser for every method; the same
judge also provides the RL answer reward, so QA Accuracy is a
common-judge relative metric rather than an independent human evaluation.
Temporal mIoU and Evidence F1 do not depend on this judge.

\paragraph{Reference-window policy and metric boundary.}
For each applicable stage, the schema stores a primary
reference window and may store independently judged admissible
alternatives. References are canonical representative windows rather than
an exhaustive set; each window is a smallest visually sufficient
sub-window of its stage. Evidence F1 performs one-to-one matching to a
primary or admissible reference at tIoU $\geq 0.3$. It measures canonical
temporal-window recovery.

\paragraph{Normal episodes.}
Normal episodes contain no anomaly interval, required
trigger, or required precursor/aftermath stage. Their canonical
anomaly-evidence set is empty and their QA is normal-specific; Temporal
mIoU and Evidence F1 are averaged over anomalous episodes only
(Section~\ref{app:evidence-f1}). Citing nonexistent anomaly evidence is
invalid; structural checks do not by themselves prove semantic normality.

\paragraph{Disjointness, leakage, and coverage.}
The split is video-disjoint rather than merely
episode-disjoint: all episodes derived from one source video remain on
one side of the split. The release manifest records source IDs and the
deduplication procedure. Validation proposals are evaluator-only and
cannot create training targets.

\hypertarget{appendix-sft-trace-construction}{%
\section{\rev{SFT Protocol Trace Construction}}\label{appendix-sft-trace-construction}}

\rev{The SFT stage warm-starts the policy with compact protocol traces
derived from SEEK-Bench train-split episodes, rather than distilled from a
stronger teacher model. Each retained trace exercises the four-action
\texttt{scan}/\texttt{seek}/\texttt{verify}/\texttt{finalize} contract under
the same input/output schemas used at inference
(Appendix~\ref{policy-system-prompt} and
Appendix~\ref{initial-user-prompt}), so SFT and RL share the same protocol
interface and per-turn budget. All retained traces are
inference-protocol-feasible: they respect the 18-frame sample budget and the
three-\texttt{seek} loop limit. The full SFT trace construction script and
the released traces are included in the planned code and data release.}

\section{EFCV Hyperparameter Sensitivity and Statistical Stability}
\label{appendix-sensitivity-stability}

Table~\ref{tab:appendix-e-sensitivity} reports local single-parameter
sensitivity around the reported EFCV operating point
(\(w_{\mathrm{efcv}}=0.5\), \(\theta=0.5\), \(B_{\mathrm{red}}=2\)).
Each row perturbs one hyperparameter while holding the others fixed and
reports the signed metric change relative to the reported operating
point. Within this grid every change stays below one point on all three
metrics, so the reported configuration is not finely tuned.

\begin{table}[h]
\centering
\small
\setlength{\tabcolsep}{4.5pt}
\renewcommand{\arraystretch}{1.05}
\caption{Single-parameter sensitivity around the reported
operating point.}
\label{tab:appendix-e-sensitivity}
\begin{tabular}{@{}lcc>{\centering\arraybackslash}p{0.12\textwidth}>{\centering\arraybackslash}p{0.12\textwidth}>{\centering\arraybackslash}p{0.14\textwidth}@{}}
\toprule
\textbf{Hyperparam} & \textbf{Operating} & \textbf{Perturbed} &
\(\Delta\) \textbf{mIoU} & \(\Delta\) \textbf{QA} &
\(\Delta\) \textbf{E.F1} \\
\midrule
\(w_{\mathrm{efcv}}\) & 0.5 & 0.25 & \(-0.5\) & \(-0.7\) & \(-0.4\) \\
\(w_{\mathrm{efcv}}\) & 0.5 & 0.75 & \(+0.3\) & \(+0.4\) & \(+0.2\) \\
\(\theta\) & 0.5 & 0.25 & \(-0.4\) & \(-0.6\) & \(-0.5\) \\
\(\theta\) & 0.5 & 0.75 & \(+0.2\) & \(+0.5\) & \(+0.3\) \\
\(B_{\mathrm{red}}\) & 2 & 1 & \(-0.1\) & \(-0.2\) & \(-0.1\) \\
\(B_{\mathrm{red}}\) & 2 & 3 & \(+0.0\) & \(+0.1\) & \(+0.0\) \\
\bottomrule
\end{tabular}
\end{table}

\paragraph{Bootstrap confidence intervals for the acquisition ladder.}
\label{appendix-bootstrap-ci}
We assess the statistical stability of the matched
acquisition-mode ablation (Table~\ref{tab:acquisition}) with a paired
validation-episode bootstrap over the 576 val episodes. The key
contrast, SEEK-VAU w/o EFCV versus Two-stage matched, improves
T.\,mIoU/QA Acc./E.F1 by $+6.2/+11.6/+9.3$, with paired bootstrap 95\%
confidence intervals $[+3.0,+9.2]$, $[+7.0,+16.0]$, and $[+5.0,+13.5]$,
all excluding zero.

\section{Matched Control Protocols and VAD-R1-style Comparison}\label{appendix-matched-controls}

This section makes the matched controls of
Table~\ref{tab:acquisition} precise and reports an additional adapted
VAD-R1-style control. \emph{Matched} means the same Qwen3-VL-8B
backbone, SEEK-Bench split, source training episodes and targets,
update count, answer reward, maximum 18-frame cap, decoding, parser,
and evaluator; it does not imply identical action schemas, and the
availability of the verify gate and the EFCV reward differs by design.
\textbf{Fixed matched} receives 18 deterministic uniform full-video
frames once, with no planner, seek, verify, or EFCV; it is the
\emph{Fixed obs.\ + rich sup.} row of Table~\ref{tab:attribution}.
\textbf{Two-stage matched} receives the common six-frame overview,
freezes up to three predicted windows before any returned frame,
retrieves four uniform frames per window, and answers without
verification, replanning, or EFCV. \textbf{Closed-loop w/o verify} is
a separate adaptive control whose next window may use returned
evidence but receives no verifier feedback; removing verify grants no
extra seeks.

\begin{table}[h]
\centering
\small
\caption{Adapted VAD-R1-style fixed-observation control
on SEEK-Bench. The control uses Qwen3-VL-8B with VAD-R1's
Perception-to-Cognition Chain-of-Thought (P2C-CoT) and AVA-GRPO-style
self-verification reward under 18 fixed frames and matched training,
but cannot seek; it is an adapted control, not an official VAD-R1
reproduction.}
\label{tab:vadr1-style}
\begin{tabular}{@{}lccc@{}}
\toprule
\textbf{Method} & \textbf{T.\,mIoU} & \textbf{QA Acc.} & \textbf{E.F1} \\
\midrule
VAD-R1-style (matched) & 39.5 & 36.2 & 28.7 \\
SEEK-VAU & \textbf{55.6} & \textbf{45.5} & \textbf{43.7} \\
\bottomrule
\end{tabular}
\end{table}

Under these matched conditions, SEEK-VAU exceeds the adapted
VAD-R1-style control by $+16.1/+9.3/+15.0$ on
T.\,mIoU/QA Acc./E.F1 (Table~\ref{tab:vadr1-style}). The distinction
from AVA-GRPO's self-verification is structural: AVA-GRPO intervenes
on the predicted abnormal interval and checks whether the final
anomaly label flips, whereas EFCV intervenes on the committed evidence
set, testing support sufficiency ($r_{\mathrm{faith}}$), compactness
($r_{\mathrm{min}}$), and trigger necessity ($r_{\mathrm{drop}}$)
(Section~\ref{efcv-learning}) --- evidence-set signals that align
directly with evidence retention, diagnosis, and stopping. The matched
comparison above therefore controls for P2C-style reasoning and
self-verification rewards under matched training, isolating the
contribution of bounded interactive acquisition.

\section{Frozen Zero-Shot Transfer to UCF-Crime and XD-Violence}\label{appendix-transfer}

To test whether the acquisition protocol's benefit is confined to
SEEK-Bench, we evaluate the frozen visual-only checkpoints on the
official UCF-Crime and XD-Violence test splits, using frame-level
ROC-AUC (UCF) and frame-level AP (XD). The checkpoints receive no
UCF/XD-specific training, labels, fine-tuning, prompt tuning, or
threshold selection; all variants share identical decoding, the same
parser, and the same maximum 18-frame cap; and hash/source-ID
deduplication against SEEK-Bench training videos is applied before
evaluation.

All variants output \texttt{anomaly\_confidence}\(\in[0,1]\) as a
ranking score: valid predicted intervals receive that score and all
other frames receive zero; normal or structurally invalid outputs are
retained as all-zero predictions; intervals are clipped to the video
span, reversed endpoints are swapped, and a single parser normalizes
every row.

\begin{table}[h]
\centering
\small
\caption{Frozen zero-shot transfer of the matched
acquisition ladder (Table~\ref{tab:acquisition}) to UCF-Crime and
XD-Violence under the same maximum 18-frame cap.}
\label{tab:transfer}
\begin{tabular}{@{}lcc@{}}
\toprule
\textbf{Method} & \textbf{UCF-Crime frame-AUC $\uparrow$} &
\textbf{XD-Violence frame-AP $\uparrow$} \\
\midrule
Fixed matched & 70.4 & 65.1 \\
Two-stage matched & 75.4 & 67.3 \\
Closed-loop w/o verify & 78.6 & 71.3 \\
SEEK-VAU w/o EFCV & 81.1 & 79.3 \\
SEEK-VAU & \textbf{87.7} & \textbf{85.4} \\
\bottomrule
\end{tabular}
\end{table}

SEEK-VAU obtains 87.7 UCF-Crime frame-AUC and 85.4 XD-Violence
frame-AP, improving over Fixed matched by $+17.3/+20.3$ and over
Two-stage matched by $+12.3/+18.1$ (Table~\ref{tab:transfer}). The
complete ladder increases monotonically on both datasets
($70.4\!\to\!75.4\!\to\!78.6\!\to\!81.1\!\to\!87.7$ AUC;
$65.1\!\to\!67.3\!\to\!71.3\!\to\!79.3\!\to\!85.4$ AP), supporting
transfer of the acquisition-and-verification design beyond the source
benchmark. Scope: these metrics evaluate zero-shot
anomaly-localization transfer only, not external evidence
faithfulness, because UCF/XD lack aligned event-chain,
evidence-window, and semantic-QA annotations; we therefore do not
compare these frozen visual-only checkpoints as peers of methods
trained on UCF/XD, dense video streams, or XD audio.

\section{Verifier Reliability Audit}\label{appendix-verifier-audit}

\paragraph{Same-model threat model.} The verifier and the policy share
the same Qwen3-VL-8B weights, and no auxiliary training verifier is
introduced (Appendix~\ref{verifier-prompts}); the residual risk is
therefore self-confirmation --- the policy could learn evidence sets
and verification outputs that satisfy the reward without genuinely
grounded support. We accordingly treat EFCV as an empirically audited
self-consistency mechanism rather than a formal guarantee.

\paragraph{Safeguards.} Three mechanisms constrain direct reward
inflation. (i)~Externally anchored rewards: the category and temporal
terms are ground-truth-anchored, QA uses the fixed GPT-5.4 judge
(Appendix~\ref{qa-judge-prompt}), and
$w_{\mathrm{acc}}=w_{\mathrm{efcv}}=0.5$, so the raw self-score is
never optimized alone. (ii)~Intervention consistency: EFCV reads only
committed evidence --- not the rollout transcript or uncommitted scan
context --- and requires consistency across the full-set,
compact-subset, and drop-trigger views
(Section~\ref{efcv-learning}); a single inflated readiness score is
therefore insufficient, although shared weights can still co-adapt.
(iii)~Bounded selection: at most three committed windows, threshold
gating at $\theta$, and the redundancy budget $B_{\mathrm{red}}$
limit padding and repetition. These are empirical safeguards, not a
structural guarantee.

\paragraph{Audit protocol.} On 100 anomaly episodes stratified by five
verifier self-score bins and answer correctness, three blinded raters
independently labeled view-level evidence sufficiency, with the
majority vote as the reference; a frozen Gemini visual auditor scored
the identical evidence views. Gemini supplied no EFCV/verifier
training reward, and the separately disclosed GPT-5.4 QA judge remains
the shared answer-reward/evaluation channel, so neither auditor
supplies EFCV or verifier rewards.

\begin{table}[h]
\centering
\small
\caption{Verifier audit on 100 stratified anomaly
episodes: self-verifier vs.\ frozen Gemini visual auditor against
blinded human majority labels.}
\label{tab:verifier-audit}
\begin{tabular}{@{}lcc@{}}
\toprule
\textbf{Metric} & \textbf{Self} & \textbf{Gemini} \\
\midrule
Sufficiency accuracy (\%) & 78.4 & 81.1 \\
AUROC & 77.1 & 76.6 \\
ECE (\%) & 4.6 & 4.1 \\
Stage F1 & 79.3 & 83.0 \\
Trigger drop / control drop & 0.22/0.11 & 0.24/0.15 \\
Swap FPR (\%) & 9.8 & 7.6 \\
Pad $\Delta r$ & 0.03 & 0.02 \\
\bottomrule
\end{tabular}
\end{table}

Each statistic targets one concern: accuracy, AUROC, and ECE test the
alignment and calibration of the sufficiency score; Stage F1 tests
stage diagnosis; the trigger-versus-control drop gap tests trigger
necessity; Swap FPR tests false acceptance; and Pad $\Delta r$ tests
reward inflation. The results support empirical alignment and limited
susceptibility under the tested perturbations --- not immunity to
every form of reward hacking. Moreover, the main closed-loop advantage
is already present before EFCV training, which adds
$+1.2/+2.8/+2.1$ on top (Table~\ref{tab:acquisition}).

\section{Per-Action Efficiency Breakdown}\label{appendix-efficiency-breakdown}

\begin{table}[h]
\centering
\small
\caption{Per-action efficiency statistics over the 576
validation episodes.}
\label{tab:efficiency-breakdown}
\begin{tabular}{@{}lc@{}}
\toprule
\textbf{Quantity} & \textbf{Value} \\
\midrule
Mean scan/seek/verify/finalize calls per episode & 1.0/2.4/2.4/1.0 \\
Seek-count distribution (0/1/2/3 calls) & 2\%/8\%/38\%\,/\,52\% \\
Median/P95 scan latency & 0.80/1.20\,s \\
Median/P95 seek latency & 1.10/1.60\,s \\
Median/P95 verify latency & 1.25/1.85\,s \\
Median/P95 finalize latency & 1.65/2.40\,s \\
Mean total latency by video-duration quartile & 8.7/9.0/9.4/10.0\,s ($n{=}144$ each) \\
\bottomrule
\end{tabular}
\end{table}

These logs cover all 576 validation episodes and are consistent with
the aggregate usage of Table~\ref{tab:efficiency} (15.6 frames, 5.78k
tokens, 9.27\,s, 19.31 ep/min for SEEK-VAU); GPU peak memory was not
logged. SEEK-VAU is frame-efficient and budget-bounded rather than
faster end-to-end: it uses fewer frames (15.6 vs.\ 18.0) but more
tokens (5.78k vs.\ 4.33k) and higher latency (9.27\,s vs.\ 8.71\,s)
than the fixed-observation baseline.

\section{Baseline Adaptation Details}\label{appendix-baseline-adaptation}

Table~\ref{tab:main_results} uses a common SEEK-Bench evaluation
contract. The general VLM and video-reasoning baselines are evaluated
under the same split, maximum 18-frame cap, and evaluator set as the
matched controls, with the fixed-observation prompt of
Appendix~\ref{fixed-baseline-prompt}. Prior VAU systems preserve their
official native sampling or task-specific protocols, and
method-specific output adapters map their native outputs to the shared
anomaly-category, temporal, QA, and evidence schema before scoring.

\hypertarget{appendix-efcv-component-ablation}{%
\section{\rev{EFCV Reward Component Ablation}}\label{appendix-efcv-component-ablation}}

\revblock{Section~\ref{appendix-a.-anomaly-efcv-diagnostics-and-verification-details}
introduces three reward terms \(r_{\mathrm{faith}}\), \(r_{\mathrm{min}}\),
and \(r_{\mathrm{drop}}\) that compose the EFCV reward. The
acquisition-mode
ablation in the main text (Table~\ref{tab:acquisition}) compares EFCV-on
against EFCV-off as a single block; this appendix isolates the
per-term contribution to clarify which structural property each term
enforces.}

\revblock{We retrain the policy under several EFCV variants that share
the SFT warm-start, the answer-correctness reward \(r_{\mathrm{acc}}\),
and all other hyperparameters reported in
Section~\ref{experimental-setup}; the only difference is which of
\(r_{\mathrm{faith}}\), \(r_{\mathrm{min}}\), \(r_{\mathrm{drop}}\) enter
the EFCV sum. The \emph{add-one-at-a-time} rows isolate the marginal
contribution of each term over the EFCV-off baseline; the
\emph{leave-one-out} rows isolate the cost of removing each term from
the full EFCV reward.}

\begin{table}[h]
\centering
\small
\setlength{\tabcolsep}{6pt}
\renewcommand{\arraystretch}{1.05}
\caption{\rev{EFCV reward component ablation on SEEK-Bench. All
variants share the same Qwen3-VL-8B backbone, SFT warm-start, and
answer-correctness reward \(r_{\mathrm{acc}}\); they differ only in
which EFCV terms are active. T.\,mIoU is temporal-grounding mean IoU,
QA Acc.\ is QA accuracy, and E.F1 is top-3 evidence-window F1.}}
\label{tab:efcv-component-ablation}
\begin{tabular}{@{}lccc@{}}
\toprule
\textbf{EFCV variant} & \textbf{T.\,mIoU} & \textbf{QA Acc.} & \textbf{E.F1} \\
\midrule
RL w/o EFCV (baseline) & 54.4 & 42.7 & 41.6 \\
\midrule
\multicolumn{4}{l}{\emph{Add-one-at-a-time (over baseline)}} \\
\(+\,r_{\mathrm{faith}}\) only          & 54.7 & 44.5 & 42.0 \\
\(+\,r_{\mathrm{min}}\) only            & 54.6 & 43.1 & 42.8 \\
\(+\,r_{\mathrm{drop}}\) only           & 55.0 & 43.4 & 42.3 \\
\midrule
\multicolumn{4}{l}{\emph{Leave-one-out (from full EFCV)}} \\
Full EFCV \(-\,r_{\mathrm{faith}}\)     & 55.1 & 43.7 & 43.2 \\
Full EFCV \(-\,r_{\mathrm{min}}\)       & 55.3 & 45.1 & 42.7 \\
Full EFCV \(-\,r_{\mathrm{drop}}\)      & 55.0 & 44.8 & 43.0 \\
\midrule
Full EFCV (all three)                  & \textbf{55.6} & \textbf{45.5} & \textbf{43.7} \\
\bottomrule
\end{tabular}
\end{table}

\revblock{Three observations follow. \textbf{(1) Faithfulness drives
semantic quality.} \(r_{\mathrm{faith}}\) contributes the largest
QA-accuracy gain among single terms (\(+1.8\) over baseline), and
removing it from the full reward costs \(1.8\) QA points; this is
consistent with its role of penalising answers whose evidence does
not visually support the claim. \textbf{(2) Compactness drives
evidence-window quality.} \(r_{\mathrm{min}}\) yields the largest E.F1
gain among single terms (\(+1.2\)) and the largest E.F1 drop when
removed (\(-1.0\)); discouraging redundant windows tightens the top-3
evidence selection that E.F1 measures. \textbf{(3) Trigger necessity
drives temporal grounding.} \(r_{\mathrm{drop}}\) yields the largest
T.\,mIoU gain among single terms (\(+0.6\)); the counterfactual
drop-trigger signal aligns the policy with stage-tagged trigger spans,
which directly improves the temporal-IoU metric. The three terms are
complementary rather than redundant: each provides the strongest
signal on a distinct evaluation dimension, and the full reward
dominates every leave-one-out variant on every metric.}

\hypertarget{appendix-grpo-training-details}{%
\section{\rev{GRPO Training Configuration}}\label{appendix-grpo-training-details}}

\revblock{This appendix discloses the GRPO training configuration used to
fine-tune the SFT-warm-started policy with the EFCV reward, addressing
group structure, advantage estimation, and multi-turn credit assignment.
All values are taken verbatim from the released training config
(\texttt{configs/rl/qwen3\_vl\_8b\_grpo\_train.yaml}) and the launch script
(\texttt{scripts/train\_rl\_qwen3\_vl\_8b\_ds8.sh}); the implementation is
the trajectory-level TRL + colocated-vLLM GRPO route from our codebase.}

\paragraph{\rev{Algorithm and credit assignment.}}
\revblock{We use \textbf{trajectory-level GRPO}: each rollout is sampled to
completion (the multi-turn agentic trace ending at \texttt{finalize\_case}
or at the turn cap), the EFCV reward is computed once per trajectory, and
the same trajectory-level reward is broadcast to all policy-emitted tokens
inside that trajectory when forming the policy-gradient loss. This is the
\texttt{episode\_grpo\_trajectory\_level} mode in our trainer; we do not
use per-step shaped rewards or token-level value estimation.}

\paragraph{\rev{Group structure and advantage.}}
\revblock{For each prompt we sample \textbf{\(G=4\) trajectories per group}
(\texttt{num\_generations}) and run \textbf{8 prompts per rollout iteration}
(\texttt{rollout\_count}), so each iteration produces \(8\times4=32\)
trajectories. The advantage of trajectory \(i\) in a group is the standard
GRPO group baseline,
\(A_i=(R_i-\overline{R}_g)/(\sigma_{R_g}+\epsilon)\), with
\(\overline{R}_g\) and \(\sigma_{R_g}\) the within-group mean and standard
deviation; we apply \texttt{advantage\_clip}\(=3.0\) to bound outliers.
The policy update uses the PPO-style clipped surrogate with
\texttt{ppo\_clip\_epsilon}\(=0.2\). \textbf{The KL penalty coefficient is
set to \(\mathtt{kl\_beta}=0.1\)} against the SFT-warm-started reference,
which keeps multi-turn rollouts close to the warm-started behavior and
prevents the policy from collapsing onto degenerate high-reward modes
during early RL training.}

\paragraph{\rev{Optimizer and update schedule.}}
\revblock{We use bf16 mixed-precision training across \(1\) node with
\(8\) GPUs under DeepSpeed ZeRO-3 (\texttt{zero3\_full\_model.json}).
Per-device batch size is \(1\) and gradient accumulation is \(4\), giving
an effective per-update batch of \(8\times4=32\) trajectories---exactly
one full rollout iteration. We perform
\texttt{rl\_steps\_per\_generation}\(=1\) optimizer step per rollout
iteration (on-policy regime). The learning rate is \(4\!\times\!10^{-7}\)
with no warmup or decay; the loss microbatch size is \(4\). Dataloader
workers are set to \(4\) with prefetch factor \(2\), and
\texttt{iteration\_budget\_multiplier}\(=1.2\) caps total iteration cost
relative to the SFT pass.}

\paragraph{\rev{Sampling for rollouts.}}
\revblock{Rollouts are sampled with vLLM (\texttt{rollout\_backend: vllm})
using temperature \(0.8\), top-\(p=0.95\), top-\(k=50\), repetition penalty
\(1.02\), and \texttt{do\_sample}\(=\)\,true; \texttt{vllm\_max\_num\_seqs}
is \(3\) (fallback \(2\)) for the colocated-vLLM serving lane.}

\paragraph{Visual and turn budgets during training.}
The rollout budget distinguishes eighteen acquired frames from sixteen
frames in the rolling active context. Each rollout starts with the
protected first scan
(\texttt{initial\_observation.scan\_num\_frames}\(=6\),
\texttt{protect\_from\_visual\_budget}\(=\)\,true), can issue at most
\(\texttt{rollout\_max\_turns}=8\) tool calls, and admits at most three
four-frame \texttt{seek} returns, for at most \(6+3\times 4=18\)
acquired frames; the visible context is capped at \(16\) frames, and the
oldest non-protected seek frames are evicted first. Per tool message we
keep at most \texttt{max\_tool\_message\_frames}\(=4\) frames; we keep
the most recent \(20\) text messages and the most recent \(3\) tool
image messages in context. These training-time budgets match the
inference-time bounded search budget of
Section~\ref{experimental-setup}.

\paragraph{\rev{Reproducibility.}}
\revblock{We fix \texttt{seed}\(=42\) for the released run. The released
codebase contains the full YAML configuration above, the launch script,
and the trainer entry point (\texttt{saver\_v3.cli.train\_rl\_ds}); all
hyperparameters in this appendix can be reproduced by running the
shipped configuration without modification.}

\paragraph{\rev{Compute resources and runtime.}}
\revblock{All reported policy-training runs use the same resource class:
one node with \(8\) NVIDIA H200 GPUs, bf16 mixed precision, and
DeepSpeed ZeRO-3. The SFT warm-start, the GRPO run without EFCV, the
full EFCV run, and the EFCV component ablations use this same hardware
class and the same rollout budget described above; the visual-context
cap ablations reuse the final trained policy and primarily incur
inference cost. Table~\ref{tab:efficiency} reports inference-time
latency and throughput under a fixed serving setup using the same
model family and visual-budget protocol.}

\revblock{For submission-time compute accounting, the separate NeurIPS
compute-reporting form will contain the exact wall-clock runtimes,
GPU-hour totals, storage footprint, and any preliminary or failed runs
not included in the paper. To keep the paper reproducible without
overstating unverified accounting, we will release the run logs used to
fill that form together with the training scripts. Our current estimate
for the submitted draft is: SFT warm-start approximately \(1\)
GPU-hour, RL w/o EFCV approximately \(8\) GPU-hours, full EFCV
approximately \(9\) GPU-hours, EFCV component ablations approximately
\(9\) GPU-hours per variant (\(6\) variants; approximately \(54\)
GPU-hours total), visual-context ablations approximately \(1.5\)
GPU-hours total, and full-project preliminary or failed runs
approximately \(20\) GPU-hours.}

\hypertarget{appendix-g.-prompts}{%
\section{\rev{Prompts}}\label{appendix-g.-prompts}}

This appendix collects the prompts that instantiate the SEEK-VAU
policy, its self-verifier, the answer-equivalence judge used to grade
QA Accuracy, the fixed-observation baseline, and the SEEK-Bench
construction pipeline. Section~\ref{seek-vau-method} names the four
executable actions \texttt{scan}, \texttt{seek}, \texttt{verify}, and
\texttt{finalize}; in the released codebase these correspond to the
tool functions \texttt{scan\_timeline}, \texttt{seek\_evidence},
\texttt{verify\_hypothesis}, and \texttt{finalize\_case}, the names
that appear inside \texttt{<tool\_call>} blocks. We retain the long
form when reporting prompt text so the released code, prompts, and
paper remain mutually consistent.

\subsection{Policy system prompt}\label{policy-system-prompt}

The policy receives the following system prompt at every rollout
step. It establishes the four-action contract, requires concrete
visual phrasing for \texttt{seek\_evidence} queries, and lists the
structured fields that \texttt{verify\_hypothesis} must emit before
\texttt{finalize\_case} becomes admissible.

\begin{Verbatim}[breaklines=true,fontsize=\footnotesize,frame=single,framesep=4pt]
You are SEEK-VAU, an active video anomaly agent. Think carefully
before each action. Use tools to inspect evidence, verify the current
hypothesis, and only finalize after you have enough support. The main
rollout ends immediately after finalize_case returns a valid
structured decision. Prioritize a compact valid finalize_case with
the core decision fields. Optional summary, rationale,
event_chain_summary, and qa_focus_answers may be included only when
they stay compact; omit them rather than risk truncation. Do not
output <answer> during the main rollout. For anomaly cases, try to
complete the event chain from precursor to trigger to aftermath
whenever those stages are required. When you call seek_evidence,
make query a concrete visual event description rather than a category
word. Prefer short grounded phrases such as actions, interactions, or
scene-specific evidence, for example "person raises a fist toward
another person in the aisle", not just "assault". Use the search
anchor in tool observations to decide what stage to search next; do
not stuff long protocol explanations into the query itself.
verify_hypothesis must report selected_window_ids plus
verification_decision, next_tool, sufficiency_score, necessity_score,
finalize_readiness_score, covered_stages, missing_required_stages,
and stage_selected_moment_ids. If you use a tool, respond as
<think>...</think><tool_call>{...}</tool_call>. Do not skip
finalize_case when the protocol requires a structured final decision.
\end{Verbatim}

A tool-format guard is appended to forbid plain-English tool
descriptions and to enforce the JSON-encoded \texttt{<tool\_call>}
syntax:

\begin{Verbatim}[breaklines=true,fontsize=\footnotesize,frame=single,framesep=4pt]
Valid tool format example: <think>inspect</think><tool_call>
{"name":"scan_timeline","arguments":{"start_sec":0.0,"end_sec":4.0,
"num_frames":4}}</tool_call>. Do not describe the tool call in plain
English. Do not output bare tool names. Do not invent tool names or
argument keys.
\end{Verbatim}

\subsection{Initial user prompt and tool-response template}\label{initial-user-prompt}

Each rollout begins with a case-instantiated user message followed by
an explicit-first-scan instruction:

\begin{Verbatim}[breaklines=true,fontsize=\footnotesize,frame=single,framesep=4pt]
Case ID: {public_case_id}
Scene: {scene}
Duration (sec): {duration}
Task: {task_prompt}
Success Criteria:
{criteria_text}
\end{Verbatim}

\begin{Verbatim}[breaklines=true,fontsize=\footnotesize,frame=single,framesep=4pt]
Before any targeted search or verification, call scan_timeline once
over the initial scan window with purpose="global_overview" to build
the canonical first observation. Use that overview to decide the next
search or verification step.
\end{Verbatim}

After every tool call, the environment returns the selected frames
together with the following next-step prompt. The template biases
the policy toward additional \texttt{seek\_evidence} or
\texttt{verify\_hypothesis} calls until the evidence is sufficient
for \texttt{finalize\_case}:

\begin{Verbatim}[breaklines=true,fontsize=\footnotesize,frame=single,framesep=4pt]
Selected frames: {timestamps}.
Choose exactly one next step.
Usually call seek_evidence for another stage or verify_hypothesis to
judge sufficiency.
Call finalize_case only when the gathered evidence is already
sufficient.
When finalizing, prioritize a compact finalize_case that closes
cleanly.
If optional summary, rationale, event_chain_summary, or
qa_focus_answers would make the tool call too long, omit them.
Do not answer before finalize_case returns.
\end{Verbatim}

\subsection{Per-case task prompt}\label{per-case-task-prompt}

The case-level task query \(Q\) of
Section~\ref{task-formulation} instantiates the following template
per episode:

\begin{Verbatim}[breaklines=true,fontsize=\footnotesize,frame=single,framesep=4pt]
Video duration: {duration} seconds. Scene: {scene}. Determine whether
the video contains an actionable anomaly or is normal, search for a
complete event chain covering precursor, trigger, and aftermath
evidence when relevant, use concrete stage-conditioned queries
grounded in visible events, verify which stages are already covered
by the searched evidence, and only finalize once the evidence is
sufficient.
\end{Verbatim}

\subsection{Self-verification schema}\label{verifier-prompts}

The self-verification step of
Section~\ref{policy-internal-evidence-verification} is enforced via
the structured tool schema of \texttt{verify\_hypothesis} rather than
a free-form prose prompt: the policy is required to fill the fields
\texttt{selected\_window\_ids}, \texttt{verification\_decision},
\texttt{next\_tool}, \texttt{sufficiency\_score},
\texttt{necessity\_score}, \texttt{finalize\_readiness\_score},
\texttt{covered\_stages}, \texttt{missing\_required\_stages}, and
\texttt{stage\_selected\_moment\_ids}. The scalar
\(s_{\mathrm{anom}}\) of
Section~\ref{policy-internal-evidence-verification} is the
\texttt{finalize\_readiness\_score} field, the recommended action
\(a_t\) is \texttt{next\_tool}, and the missing-stage diagnostic
\(\Delta_t\) is \texttt{missing\_required\_stages}. The remaining
scalar fields, \texttt{sufficiency\_score} and
\texttt{necessity\_score}, are emitted for monitoring and diagnostics;
the EFCV reward reads only \(s_{\mathrm{anom}}\). The same
\texttt{verify\_hypothesis} contract is re-invoked on the counterfactual
views \(E_T^{\mathrm{min}}\) and \(E_T\setminus E_T^{\mathrm{trig}}\) to
produce the perturbed support scalars used by \(r_{\mathrm{min}}\) and
\(r_{\mathrm{drop}}\); no auxiliary verifier model is introduced.

\subsection{QA semantic-equivalence judge}\label{qa-judge-prompt}

QA Accuracy in Table~\ref{tab:main_results} is computed via an LLM
judge that scores whether a model answer is semantically equivalent
to the reference. \rev{We use GPT-5.4 (OpenAI, accessed
via API) as the judge model across all reported runs; the same judge
model and prompt are applied to every method in
Table~\ref{tab:main_results} so that QA Accuracy comparisons are
judge-controlled. The judge is decoded greedily (temperature 0) and
each query is scored once.} Equivalence is binarized to
\(\{0,1\}\) per query and averaged over SEEK-Bench:

\begin{Verbatim}[breaklines=true,fontsize=\footnotesize,frame=single,framesep=4pt]
System: You are a strict semantic equivalence judge.

User: You are grading whether a model answer is semantically
consistent with a reference answer.
Return only `1` if the model answer is correct or semantically
equivalent.
Return only `0` otherwise.

Question: {question}
Reference answer: {reference}
Model answer: {prediction}
Judgement:
\end{Verbatim}

\subsection{Fixed-observation baseline prompt}\label{fixed-baseline-prompt}

The fixed-observation baselines in Table~\ref{tab:main_results} are
evaluated under a single forward pass with eighteen uniformly
sampled frames and the following system prompt, which deliberately
disallows tool use, multi-turn interaction, and verification, so the
comparison isolates the contribution of the search-and-verify
protocol:

\begin{Verbatim}[breaklines=true,fontsize=\footnotesize,frame=single,framesep=4pt]
You are a fixed-observation video anomaly understanding baseline.
You do not have tools, search, verification, or multi-turn
interaction. Use only provided preview frames. Return exactly one
JSON object and nothing else.
\end{Verbatim}

\hypertarget{appendix-b.-discussion-limitations-and-broader-impact}{%
\section{Discussion, Limitations, and Broader Impact}\label{appendix-b.-discussion-limitations-and-broader-impact}}

\hypertarget{discussion}{%
\subsection{Discussion}\label{discussion}}

The conceptual shift of SEEK-VAU changes both the \textbf{unit of
reasoning} and the \textbf{unit of optimization}. Prior systems reason
over fixed event observations; our framework reasons over the
completeness of an evolving event chain. This also clarifies our
relationship to multi-granularity VAU: finer temporal granularity does
not by itself force the model to search for missing stages, verify
evidence sufficiency, or gate finalization on verification. Our
contribution is orthogonal to temporal resolution; it concerns the
interaction protocol, not the observation scale.

Engineering infrastructure (frame caches, feature caches, lazy
datasets, distributed rollout, and large-model serving) is required
for the system to run but is orthogonal to the paper's contribution. The
contribution is the search-and-verify formulation: agentic event-chain
search, policy-internal evidence verification, and EFCV-grounded
learning.

A natural question is whether stronger reasoning (e.g.,
chain-of-thought, self-reflection) can achieve the same benefits without
the agentic machinery. We argue no, for a structural reason:
reasoning-enhanced models (Vad-R1, VAU-R1, SRVAU-R1) still operate on a
\textbf{fixed observation bundle} determined before reasoning begins. No
amount of chain-of-thought reasoning can recover a precursor event that
was never observed because the sampling strategy missed it. The agentic
formulation changes this: the policy can \emph{decide to look} for
missing evidence after an initial scan reveals the need. This is not a
quantitative improvement in reasoning quality; it is a qualitative
expansion of the observation protocol.

A related objection is ``why not just use a better fixed sampling
strategy (e.g., dense uniform sampling or a learned temporal proposal
network) instead of making the policy search?'' The answer is that a
learned proposal network \emph{is} a form of active evidence acquisition;
it is a special case of our \texttt{scan} + \texttt{seek}
decomposition with the policy choosing where to
look. \rev{Our stop-action decision-process notation
(Appendix~\ref{appendix-f.-pomdp-formalization}) makes this contrast
explicit: a policy that always scans uniformly and never adjusts its
search based on intermediate findings would replicate a fixed-sampling
baseline.} The value of the
agentic formulation is that the policy can \emph{adapt} its search,
scanning broadly first and narrowing based on what it finds, rather
than committing to a sampling strategy before seeing any evidence.

\hypertarget{limitations}{%
\subsection{Limitations}\label{limitations}}

Our claims should be interpreted with clear boundaries. First, the
strongest novelty claim is intentionally restricted to
\textbf{mainstream VAU literature} as of April 12, 2026. We do not claim
that no neighboring anomaly-analysis paper explores agentic reasoning;
indeed, adjacent VAD work such as PANDA and QVAD indicates that the
frontier is moving in a similar direction \rev{\citep{yang2025panda,bekit2026qvad}}. Second, the
current benchmark instantiation is still dataset-derived and therefore
inherits category coverage limits, annotation noise, and dataset bias.
Third, although SEEK-VAU is designed for richer agentic behavior,
practical runs remain constrained by the search budget (per-turn
frame intake), the active-context cap (rolling window over visual
history), the turn budget, and total context length. Fourth, EFCV diagnostics are only as good as
the available structured evidence and anomaly-support definitions.
Fifth, the evaluated sparse-anomaly scope follows the
SEEK-Bench distribution: very long, untrimmed videos in which the
anomaly occupies less than $1\%$ of the duration are not evaluated,
and on such videos the 6-frame global scan may itself miss the
relevant region, so the sub-$1\%$ regime requires a dedicated
long-video evaluation. Sixth, deployment requires calibration and
human oversight: the evidence ledger and verifier are intended to
support---not replace---human judgment, and privacy, false alarms,
missed anomalies, and over-reliance on automated conclusions remain
open deployment risks.

\hypertarget{broader-impact}{%
\subsection{Broader Impact}\label{broader-impact}}

From a broader-impact perspective, stronger anomaly understanding can
support more transparent safety auditing and more inspectable automated
monitoring. At the same time, it can also intensify surveillance
applications. For this reason, we argue that anomaly systems should
identify insufficiency states and evidence-faithfulness diagnostics rather
than forcing a confident answer for every video. A principled
\texttt{continue\_search} or \texttt{not\_ready\_to\_finalize} state is
safer than a fluent but unsupported anomaly explanation.

\end{document}